\documentclass{article}

\usepackage{microtype}
\usepackage{graphicx}
\usepackage{booktabs}
\usepackage{hyperref}

\usepackage{xcolor}
\usepackage[accepted]{icml2021}

\icmltitlerunning{Composing Normalizing Flows for Inverse Problems}

\usepackage{amsmath,amsfonts,bm}

\def\eqref#1{equation~\ref{#1}}
\def\1{\bm{1}}

\def\rmI{{\mathbf{I}}}

\def\vy{{\bm{y}}}

\DeclareMathAlphabet{\mathsfit}{\encodingdefault}{\sfdefault}{m}{sl}
\SetMathAlphabet{\mathsfit}{bold}{\encodingdefault}{\sfdefault}{bx}{n}

\def\gL{{\mathcal{L}}}

\def\gN{{\mathcal{N}}}

\def\gQ{{\mathcal{Q}}}

\def\sR{{\mathbb{R}}}

\newcommand{\E}{\mathbb{E}}

\newcommand{\KL}{D_{\mathrm{KL}}}

\DeclareMathOperator*{\argmax}{arg\,max}
\DeclareMathOperator*{\argmin}{arg\,min}

\usepackage{url}
\usepackage{microtype}
\usepackage{graphicx}
\usepackage{subcaption}
\usepackage{booktabs}
\usepackage[utf8]{inputenc}
\usepackage[T1]{fontenc}
\usepackage{hyperref}
\usepackage{url}
\usepackage{booktabs}
\usepackage{amsfonts}
\usepackage{nicefrac}
\usepackage{microtype}
\usepackage{graphicx}
\usepackage{amsmath}
\usepackage{amsthm}
\usepackage{amssymb}
\usepackage{bbm}
\usepackage{cleveref}
\usepackage{mathtools}
\usepackage{tikz}
\usetikzlibrary{arrows}
\usetikzlibrary{fit,positioning}
\usetikzlibrary{arrows.meta}
\usetikzlibrary{shapes.geometric}
\usepackage{enumitem}
\usepackage{wrapfig}

\begin{document}
\newtheorem*{rep@theorem}{\rep@title}
\newcommand{\newreptheorem}[2]{%
\newenvironment{rep#1}[1]{%
 \def\rep@title{#2 \ref{##1}}%
 \begin{rep@theorem}}%
 {\end{rep@theorem}}}
\newtheorem{theorem}{Theorem}
\newreptheorem{theorem}{Theorem}
\newtheorem{lemma}[theorem]{Lemma}
\newreptheorem{lemma}{Lemma}
\newtheorem{corollary}[theorem]{Corollary}
\newtheorem{proposition}[theorem]{Proposition}
\newreptheorem{proposition}{Proposition}
\theoremstyle{definition}
\newtheorem{observation}[theorem]{Observation}
\newtheorem{definition}[theorem]{Definition}

\newcommand{\setcomp}[1]{\overline{#1}}
\DeclarePairedDelimiterX{\infdivx}[2]{(}{)}{%
  #1\;\delimsize\|\;#2%
}
\newcommand{\kldiv}{\KL\infdivx}

\newcommand{\brac}[1]{\left[#1\right]}
\newcommand{\paren}[1]{\left(#1\right)}
\newcommand{\stackeq}[1]{\stackrel{#1}{=}}
\newcommand{\norm}[1]{\left\|#1\right\|}
\newcommand{\indi}[1]{\mathbbm{1}\brac{#1}}
\newcommand{\inner}[2]{\left\langle#1,\;#2\right\rangle}

\newcommand{\bx}{\bm{x}}
\newcommand{\bz}{\bm{z}}
\newcommand{\xobs}{\bm{x}_1}
\newcommand{\xhat}{\hat{\bx}}
\newcommand{\xmmse}{\xhat_{\text{MMSE}}}
\newcommand{\xstar}{\bx^*}
\newcommand{\xmis}{\bm{x}_2}
\newcommand{\xtil}{\tilde{\bm{x}}_1}
\newcommand{\by}{\bm{y}}
\newcommand{\ytil}{\tilde{\bm{y}}}
\newcommand{\ystar}{\vy^*}
\newcommand{\bzero}{\bm{0}}
\newcommand{\eps}{\bm{\epsilon}}
\newcommand{\Lours}{\gL_{\text{ours}}}
\newcommand{\Lnaive}{\gL_{\text{naive}}}

% Distributions
\newcommand{\fbase}{f}
\newcommand{\fcond}{\hat{f}}
\newcommand{\fambi}{g}
\newcommand{\psig}{p_{\sigma}}
\newcommand{\pbase}{p_{\bx}}
\newcommand{\pcomp}{q_{\bx}}
\newcommand{\pcond}{q_{\bz}}
\newcommand{\pambi}{p_{\fambi}}
\newcommand{\Normal}{\gN(\bzero, \rmI)}
\newcommand{\pnormal}{p_{\bz}}
\newcommand{\qnormal}{q_{\eps}}

\twocolumn[
\icmltitle{Composing Normalizing Flows for Inverse Problems}
\icmlsetsymbol{equal}{*}

\begin{icmlauthorlist}
\icmlauthor{Jay Whang}{utcs}
\icmlauthor{Erik M. Lindgren}{google}
\icmlauthor{Alexandros G. Dimakis}{utece}
\end{icmlauthorlist}
\icmlaffiliation{utcs}{Dept. of Computer Science, UT Austin, TX, USA}
\icmlaffiliation{utece}{Dept. of Electrical and Computer Engineering, UT Austin, TX, USA}
\icmlaffiliation{google}{Google Research, NY, USA}
\icmlcorrespondingauthor{Jay Whang}{\texttt{jaywhang@utexas.edu}}
\icmlkeywords{normalizing flow, inverse problems, conditional inference}

\vskip 0.3in
]

\printAffiliationsAndNotice{}

%-------------------------------------------------------------------------------%
\begin{abstract}
Given an inverse problem with a normalizing flow prior, we wish to estimate the distribution of the underlying signal conditioned on the observations.  We approach this problem as a task of conditional inference on the pre-trained unconditional flow model.
We first establish that this is computationally hard for a large class of flow models.  Motivated by this, we propose a framework for approximate inference that estimates the target conditional as a composition of two flow models.  This formulation leads to a stable variational inference training procedure that avoids adversarial training. Our method is evaluated on a variety of inverse problems and is shown to produce high-quality samples with uncertainty quantification. We further demonstrate that our approach can be amortized for zero-shot inference.
\end{abstract}

%-------------------------------------------------------------------------------%
\section{Introduction}

We are interested in solving inverse problems using a pre-trained normalizing flow prior. 
Inverse problems encompass a variety of tasks such as image inpainting, super-resolution and compressed sensing from linear projections. Due to this generality, the applications range from scientific and medical imaging to computational photography~\cite{ongie2020deep}. 
Inverse problems can be solved by either supervised~\cite{pathak2016context, richardson2020encoding, yu2018generative} or unsupervised~\cite{pulse, bora2017compressed, pajot2018unsupervised} methods, see the recent survey~\cite{ongie2020deep} for a unified presentation. 

In this paper we focus on unsupervised image reconstruction techniques that benefit from a pre-trained deep generative prior, specifically normalizing flows. Flow models \citep{papamakarios2019survey} are a family of generative models that provide efficient sampling, likelihood evaluation, and inversion.  While other types of models can outperform flow models in terms of likelihood or sample quality, flow models are often simpler to train and evaluate compared to other models.

These characteristics make normalizing flows attractive for numerous downstream tasks, including density estimation, inverse problems, semi-supervised learning, reinforcement learning, and audio synthesis \citep{ho2019flow++,asim2019invertible,whang2020compressed,atanov2019semi,ward2019improving,oord2018parallel}.

Even with such computational flexibility, how to perform efficient probabilistic inference on a flow model subject to observations obtained from an inverse problem remains challenging.  This question is becoming increasingly important as flow models increase in size, and the computational resources necessary to train them from scratch are out of reach for many researchers and practitioners\footnote{For example, \citet{kingma2018glow} report that their largest model had 200M parameters and was trained on 40 GPUs for a week.}. 
Our goal is to \textit{re-purpose} these powerful pre-trained models for different custom inverse problems without re-training them from scratch.

Concretely, we wish to recover the distribution of the unknown image $\bx$ from the observed measurements ${\ystar= A(\bx) + \text{noise}}.$ 
We assume that a pre-trained flow model $p(\bx)$ serves as the prior for natural images we are sensing, and that the measurement function $A(\cdot)$ (also known as forward operator) is differentiable. 
Thus the goal is to estimate the following conditional distribution as accurately as possible: 
$$p(\bx \mid A(\bx) = \ystar).$$

We propose a novel formulation that \textit{composes} a new flow model with the pre-trained prior $p(\bx)$ to estimate the conditional distribution given observations $\ystar$.
While such a composed model is in general intractable to train for latent variable models, the invertibility of the given prior leads to a tractable and stable training procedure via variational inference (VI).

\textbf{Our contributions:}
\begin{itemize}[topsep=1pt,itemsep=4pt,partopsep=0pt,parsep=0pt,leftmargin=5mm]
    \item We show that even though flow models are designed to provide efficient inversion and sampling, even \textit{approximately} sampling from the exact conditional distribution is computationally intractable for a wide class of flow models.  Motivated by this, we consider the relaxation that allows approximate \textit{conditioning}.
    \item We propose to estimate the relaxed target conditional distribution by composing a new flow model (the \textit{pre-generator}) with the base model. This formulation leads to a variational inference training procedure that avoids the need for unstable adversarial training as explored in existing work \citep{engel2017latent}.  
    \item Because our method recovers the conditional distribution over $\bx$ as another flow model, we can use it to efficiently generate conditional samples and evaluate conditional likelihood. This is in contrast to prior work that uses flow models for inverse problems~\cite{asim2019invertible}, since we can obtain confidence bounds for each reconstructed pixel beyond point estimates. 
    \item We show that our approach is comparable to MCMC baselines in terms of sample quality metrics such as Frechet Inception Distance (FID) \citep{heusel2017gans}. We also qualitatively demonstrate its flexibility on various complex inference tasks with applications to inverse problems.
    \item We show that the pre-generator can be \textit{amortized} over the observations to perform zero-shot inference without much degradation in sample quality.
\end{itemize}

\begin{figure}[!h]
\centering
\includegraphics[width=0.95\linewidth]{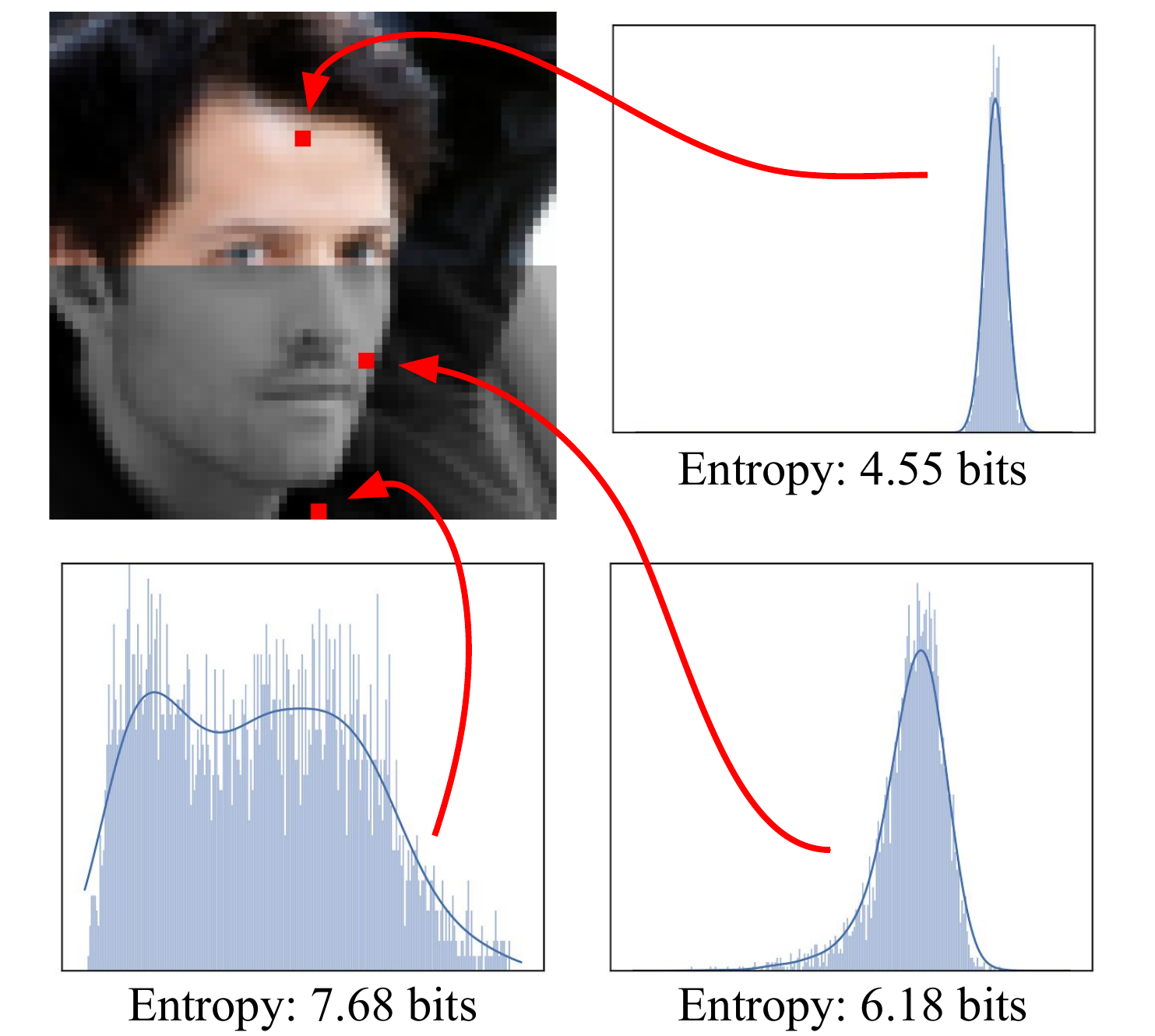}
\caption{Uncertainty quantification highlighted at different pixel locations, obtained from our learned approximate posterior.  The top pixel is observed, and thus is sharply concentrated on a single value (the small variance is due to our use of smoothing).  The bimodal plot distribution in the bottom left captures the semantic ambiguity of the bottom pixel that can be part of either the neck or the background.}
\label{fig:pixel_hist}
\end{figure}

%-------------------------------------------------------------------------------%
\section{Background}

\subsection{Normalizing Flows}
\label{sec:normalizing_flow}
Normalizing flow models represent complex probability distributions by transforming a simple input noise $\bz$ (typically standard Gaussian) through a differentiable bijection $f: \sR^d \to \sR^d$.
Since $f$ is invertible, we can compute the probability density of $\bx=f(\bz)$ via the change of variables formula: 
$\log p(\bx) = \log p(\bz) + \log \left \vert \det \frac{df^{-1}}{d \bx}(\bx) \right \vert$.

Flow models are explicitly designed so that this expression can be easily computed. This allows them to be directly trained with maximum likelihood objective on data and avoids issues such as posterior and mode collapse that plague other deep generative models.

Starting from the early works of \citet{dinh2014nice} and \citet{rezende2015variational}, there has been extensive research on invertible neural network architectures for normalizing flow. Many of them work by composing a series of invertible layers, such as in RealNVP \citep{dinh2016density}, IAF \citep{kingma2016improved}, Glow \citep{kingma2018glow}, invertible ResNet \citep{behrmann2018invertible}, and Neural Spline Flows \citep{durkan2019neural}. 

One of the simplest invertible layer construction is \textit{additive coupling layer} introduced by \citet{dinh2014nice}, which served as the basis for many other subsequently proposed models.
In an additive coupling layer, the input variable is partitioned as $\bx = (\bx_1, \bx_2) \in \sR^{d_1} \times \sR^{d_2}$. The layer is parametrized by a neural network $g(\bx_1): \sR^{d_1} \to \sR^{d_2}$ used to additively transform $\bx_2$. Thus the layer's output $\by = (\by_1, \by_2) \in \sR^{d_1} \times \sR^{d_2}$ and its inverse can be computed as follows:
$$
\begin{cases}
\by_1 &= \bx_1 \\
\by_2 &= \bx_2 + g(\bx_1)
\end{cases}
\Longleftrightarrow
\begin{cases}
\bx_1 &= \by_1 \\
\bx_2 &= \by_2 - g(\by_1)
\end{cases}
$$
Notably, the determinant of the Jacobian of this transformation is always 1 for any mapping $g$.

\subsection{Variational Inference}

\label{sec:variational_inference}
Variational inference (VI) is a family of techniques for estimating the conditional distribution of unobserved variables given the observed ones \citep{jordan1999introduction, wainwright2008graphical, blei2017variational}. At its core, VI tries to find a tractable approximation of the intractable target posterior by solving a KL minimization problem.

Within our context of conditional inference on a joint distribution $p(\bx)$, we minimize the following stochastic variational inference (SVI) objective:
\begin{equation}
\label{eqn:vi_loss}
\min_{q \in \gQ} \kldiv{q(\bx)}{p(\bx \mid \by = \ystar)},
\end{equation}
where $\by \triangleq A(\bx)$ is the measurement (also called observation) that is being conditioned on, and $\ystar$ is the given realization of $\by$. The variational family $\gQ$ must be appropriately chosen to allow efficient sampling and likelihood evaluation for all $q \in \gQ$.
Note that $q$ is specific to the particular value of $\ystar$.

The variational posterior $q$ can also be \textit{amortized} over the observation \cite{kingma2013auto}, leading to a single model trained to minimize the following amortized variational inference (AVI) objective:
\begin{equation}
\label{eqn:avi_loss}
\min_{q \in \gQ} \E_{\by}\brac{\kldiv{q(\bx \mid \by)}{p(\bx \mid \by)}},
\end{equation}
An amortized posterior has the advantage that it only needs to be trained once for all $\by$, but it generally achieves worse likelihood than SVI and often requires a more complex model \citep{cremer2018inference}.

%-------------------------------------------------------------------------------%
\section{Hardness of Conditional Sampling}
\label{sec:hardness}

Before we present our method, we first establish a hardness result for conditional sampling for flow models.  Specifically, if an algorithm is able to efficiently sample from the conditional distribution of a flow model with additive coupling layers \cite{dinh2014nice}, then it can be used to solve NP-complete problems efficiently. 
The formal statement and the proof of the theorem can be found in \Cref{appendix:hardness_results}.

\begin{theorem}
\label{thm:hardness_exact}
(Informal) Suppose we are given a flow model with additive coupling layers and wish to condition on a subset of the input dimensions. If there is an efficient algorithm that can sample from this conditional distribution, then $RP = NP$.
Further, this problem remains hard even if we allow sampling to be approximate.
\end{theorem}

Importantly, this result shows that allowing approximate \textit{sampling} from the exact posterior does not affect the hardness of the problem, as long as we require that the conditioning is exact. Thus we are motivated to consider approximate \textit{conditioning}, where the conditioned variable is allowed to deviate from the given observation.

We also note that flow architectures that include additive coupling layers make up a majority of existing models (e.g. most of the models in \Cref{sec:normalizing_flow}). Thus our hardness result applies to a variety of flow models used in practice.

%-------------------------------------------------------------------------------%
\section{Approximate Conditional Inference with Composed Flows}
\label{sec:our_method}
\begin{figure}
\centering
\begin{tikzpicture}[->,>=stealth',shorten >=1pt,auto,node distance=1.8cm,
                    thick,
                    var node/.style={circle,draw,font=\sffamily\bfseries, minimum size=0.8cm},
                    func node/.style={rectangle, draw,font=\sffamily\bfseries, minimum size=0.8cm}]
  \node[var node] (z0) {$\eps$};
  \node[func node] (fhat) [right of=z0]{$\hat{f}$};
  \node[var node] (z1) [right of=fhat]{$\bz$};
  \node[func node] (f) [right of=z1]{$\fbase$};
  \node[var node] (xy) [right of=f] {$\bx$};
  \path[every node/.style={font=\sffamily\small}]
    (z0) edge (fhat)
    (fhat) edge (z1)
    (z1) edge (f)
    (f) edge (xy);
\end{tikzpicture}
\caption{A flow chart of our conditional sampler. Gaussian noise $\eps \sim \Normal$ is mapped through the composition of our pre-generator $\fcond$ and the base model $\fbase$ to generate conditional samples.}
\label{fig:composed_flow}
\end{figure}
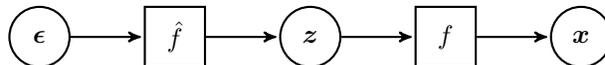

\textbf{Notation.}
Let $\pbase(\bx)$ be the pre-trained \textit{base model} that serves as the signal prior, parametrized by the invertible mapping $\fbase: \bz \mapsto \bx$. $A(\bx)$ is the differentiable measurement function.
We similarly define the \textit{pre-generator} $\pcond(\bz)$ parametrized by the mapping $\fcond: \eps \mapsto \bz$, which represents a distribution in the latent space of the base model.
We assume that all flow models use the standard Gaussian prior, i.e. $\pnormal(\bz)$ and $\qnormal(\eps)$ are $\Normal$.

By composing the base model and the pre-generator, we obtain the \textit{composed model}, denoted $\pcomp(\bx)$, whose samples are generated via $\eps \sim \Normal \to \bx = \fbase(\fcond(\eps))$.
\Cref{fig:composed_flow} details this sampling procedure.

\textbf{VI objective and smoothing.}

Since our composed model $\pcomp$ is the composition of two flow models, the VI objective in \cref{eqn:vi_loss} can be simplified further (see \Cref{appendix:loss_simplification} for derivation):
\begin{equation}
\label{eqn:latent_vi_loss_exact}
\min_{\fcond} \kldiv{\pcond}{\pnormal} + \E_{\pcond}\brac{-\log p(\by = \ystar \mid \bz)}
\end{equation}

Unfortunately this loss is challenging to optimize in practice when using a flow-based variational posterior. Because $\by = A(\fbase(\bz))$ is a deterministic function of $\bz$, the density in the second term is zero for any $\bz$ that fails to match $\ystar$ exactly. Since our pre-generator $\pcond$ is a flow model defined by an invertible mapping $\fcond$ and has full support, it would inadvertently have nonzero probability mass on invalid values of $\bz$ and cause the loss to be infinity.

One simple solution to this issue is \textit{smoothing} the observation, which turns the condition $\by = \ystar$ into a soft constraint.  Notice that this is in line with the hardness result in \Cref{sec:hardness}, where we motivated the need for approximate conditioning. 
In the context of inverse problems, smoothing can also be viewed as the distribution for observation noise.

Concretely, we define a new random variable $\ytil$ that is allowed to deviate from $\by$ but penalized for the deviation.
While there are many choices for the distribution $p(\ytil\mid\by)$, we consider the following scheme.
For any symmetric distance measure $d(\cdot,\cdot)$ with $d(\by,\ytil) = 0$ iff $\by = \ytil$, we use the distribution defined by $p(\ytil\mid\by) \propto \exp(-\beta \cdot d(\ytil,\by))$. Note that we do not need to compute the normalization constant as it is constant w.r.t. $\fcond$, which we optimize.

This formulation includes a wide range of options for smoothing.  For example, choosing $\ell_2$ distance and ${\beta = 1/(2\sigma^2)}$ is equivalent to smoothing with Gaussian kernel $\gN(\bzero, \sigma^2 \rmI)$, which leads to the following objective:
\begin{equation}
\label{eqn:latent_vi_loss_smooth}
\begin{split}
&\Lours(\fcond)
=\kldiv{\pcomp}{\pbase(\cdot \mid \ytil=\ystar)} \\
&=\kldiv{\pcond}{\pnormal} + \E_{\pcond} \brac{ \frac{1}{2\sigma^2}\norm{A(\fbase(\bz))- \ystar }_2^2}
\end{split}
\end{equation}

This loss function offers an intuitive interpretation. The first term tries to keep the learned distribution $\pcomp$ close to the base distribution by pushing $\pcond$ to match the prior of the base model, while the second term tries to match the observation $\ystar$. This is analogous to the KL/reconstruction loss decomposition typically used in the VAE literature.

We could also choose to use a more sophisticated distance measure such as LPIPS \citep{zhang2018unreasonable}. Interestingly, our preliminary experiments showed no benefit in sample quality when using LPIPS, so we ran our experiments with Gaussian smoothing for simplicity.  We leave a detailed study on the effect of different smoothing techniques for future work.

\textbf{Bounding the marginal objective.}
An important related task is estimating the marginal distribution after conditioning.  In other words, can we estimate $p(\bx_2 \mid \ytil = \ystar)$ for some partitioning of the input $\bx = (\bx_1, \bx_2)$?  This includes tasks such as data imputation, e.g. estimating $p(\bx_2 \mid \bx_1)$.

In our setup, computing $p(\bx_2 \mid \ytil = \ystar)$ is in general intractable because we only have access to the joint distribution $\pbase(\bx_1, \bx_2)$ through the base model. Fortunately, our VI loss for the joint conditional $\pbase(\bx \mid \ytil = \ystar)$ provides an upper bound (derivation in \Cref{appendix:loss_bound}):
\begin{align*}
    \text{(Joint KL)}
    &= \kldiv{\pcomp(\bx)}{\pbase(\bx \mid \ytil = \ystar)} \\
    &\ge \kldiv{\pcomp(\xmis)}{\pbase(\xmis \mid \ytil = \ystar)}.
\end{align*}
Thus we are justified in our use of \cref{eqn:latent_vi_loss_smooth} in place of the intractable marginal KL. 

\textbf{Benefits of solving inverse problems distributionally.}
Here we explain a key benefit of recovering the conditional distribution instead of just a point estimate. Given the observation $\ystar$ generated from the underlying signal $\xstar$, suppose we wish to recover $\xstar$ with respect to the $\ell_2$ loss.
Then the optimal recovery function is the minimum mean square error (MMSE) estimator
$\xmmse(\ystar) = \argmin_{\xhat} \norm{\xstar - \xhat(\ystar)}_2^2$.
Under a mild assumption, the MMSE estimator is known to be the conditional expectation:
$
\xmmse(\ystar) = \E \brac{\bx \mid \ystar}.
$

Importantly, this is different from the objective employed by existing methods that produce point estimates. For example, \citet{bora2017compressed} minimize the reconstruction error based on a projection to the range of a GAN:
$$
\xhat_{\text{bora}}(\ystar) = \argmin_{\bx \in \text{range}(G)} \norm{A(\bx) - \ystar}_2^2,
$$
and \citet{asim2019invertible} use an objective loosely based on a MAP estimate: 
$$
\xhat_{\text{asim}}(\ystar) = \argmax_{\bx} p(\bx\mid\ystar).
$$
The issue with these objectives is that, even if these optimizations could be done perfectly, they would not produce $\xmmse(\ystar)$ and thus lead to suboptimal recovery with respect to the $\ell_2$ loss.

Instead, our approach is to recover the entire conditional distribution $p(\bx | \ystar)$ and use it to obtain a Monte Carlo estimate of the conditional mean $\E\brac{\bx\mid\ystar}$. While MCMC methods can also be used for this purpose, they often take prohibitively long due to slow mixing and may produce correlated samples. Our approximate posterior is explicitly parametrized as a flow and can efficiently generate i.i.d. samples. As we will see in our experiments later, this has a significant performance benefit compared to the existing approaches in terms of reconstruction error and the speed of inference.

%-------------------------------------------------------------------------------%
\section{Related Work}

\textbf{Conditional generative models.}
There has been a large amount of work on conditional generative modeling, with varying levels of flexibility for what can be conditioned on.  In the simplest case, a fixed set of observed variables can be directly fed into the model as an auxiliary conditioning input \citep{mirza2014conditional,sohn2015learning,ardizzone2019guided}. Some recent works proposed to extend existing models to support conditioning on \textit{arbitrary} subsets of variables \citep{ivanov2018variational,belghazi2019learning,li2019flow}. This is a much harder task as there are exponentially many subsets of variables that can be conditioned on.

More relevant to our setting is \citep{engel2017latent}, which studied conditional sampling from \textit{non-invertible} latent variable generators such as VAE and GAN. It proposes to adversarially train a pre-generator, thereby avoiding the issue of intractability of VI for non-invertible models.  Due to the adversarial training and the lack of invertibility of the base model, however, the learned conditional sampler lacks the computational flexibility of a flow-based posterior, such as tractable likelihood computation and inversion.  The key difference of our method is that by explicitly parametrizing the conditional generator to be invertible as a composition of two flow models, we avoid the need for adversarial training.

We highlight several reasons why one might prefer our approach over the above methods:
(1) the training data for the base model is not available, and only the model itself is made public (2) the conditional posterior is too costly to train from scratch (3) we wish to perform downstream tasks that require exact likelihood or inversion (4) we want to get some insight on the distribution defined by the given model.

\textbf{Markov Chain Monte Carlo methods.}
When one is only concerned with generating samples, MCMC techniques offer a promising alternative.  Unlike VI using an approximate posterior, MCMC methods come with asymptotic guarantees to generate samples from the target posterior . Though directly applying MCMC methods on complex high-dimensional posteriors parametrized by a neural network often comes with many challenges in practice \citep{papamarkou2019challenges}, methods based on Langevin Monte Carlo have recently shown promising results \citep{neal2011mcmc,welling2011bayesian,song2019generative}.

The idea of leveraging the favorable geometry of the latent space of a flow model is also applicable to MCMC methods. For example, \citet{hoffman2019neutra} utilized the latent space of a flow model to improve mixing of Hamiltonian Monte Carlo. More recently \citet{cannella2020projected} proposed PL-MCMC, a Metropolis-Hastings based sampler with transition kernel also defined in the latent space of a pre-trained flow. A similar idea was later adapted by \citet{nijkamp2020learning} in the context of training energy-based models.

\begin{table*}[!ht]
\centering
\caption{Sample quality metrics for image inpainting tasks on different datasets. The best value is bolded for each metric. As shown below, our method achieves superior sample quality to all baselines.} 
\label{table:fid}
\begin{tabular}{@{}lccccccccccc@{}}
\toprule

& \multicolumn{3}{c}{MNIST} & \multicolumn{4}{c}{CIFAR-10 (5-bit)} & \multicolumn{4}{c}{CelebA-HQ (5-bit)} \\ 
\cmidrule{2-4} \cmidrule(lr){5-8} \cmidrule{9-12}

& FID & MSE & LPIPS & FID & IS $\uparrow$
& MSE & LPIPS & FID & MSE & LPIPS & Diversity $\uparrow$ \\
\midrule

Ours (SVI)      & \textbf{5.15}     & \textbf{22.13}    & \textbf{0.076}    &
\textbf{45.01}  & \textbf{7.14}     & 9.73              & \textbf{0.177}    &
37.24           & \textbf{219.7}    & \textbf{0.207}    & 0.450 $\pm$ 0.086 \\
\addlinespace[0.3em]

LMC             & 14.56             & 36.47             & 0.135             &
47.53           & 6.73              & \textbf{9.31}     & 0.201             &
\textbf{30.34}  & 323.5             & 0.229             & 0.479 $\pm$ 0.077 \\
\addlinespace[0.3em]

Ambient VI      & 123.6             & 59.99             & 0.282             &
87.50           & 5.14              & 16.59             & 0.295             &
295.0           & 2084              & 0.738             & \textbf{0.586 $\pm$ 0.186} \\
\cmidrule(lr){5-8} \cmidrule{9-12}

PL-MCMC         & 21.20  & 59.89   & 0.190  & \multicolumn{4}{c}{N/A}       & \multicolumn{4}{c}{N/A}       \\
\bottomrule

\end{tabular}
\end{table*}

\textbf{Inverse problems with deep generative prior.}
In a linear inverse problem, a vector $\bx \in \sR^d$ generates a set of measurements $\by^* = A\bx \in \sR^m$, where the number of measurements is much smaller than the dimension: $m \ll d$. The goal is to reconstruct the vector $\bx$ from $\by^*$. While in general there are (potentially infinitely) many possible values of $\bx$ that agree with the given measurements, it is possible to identify a unique solution when there is an additional structural assumption on $\bx$.

Classically, the simplifying structure was that $\bx$ is sparse  \citep{tibshirani1996regression, candes2006stable, donoho2006compressed, bickel2009simultaneous, baraniuk2007compressive}.
Recent work has considered alternative structures, such as the vector $\bx$ coming from a deep generative model. Starting with \citet{bora2017compressed}, there has been extensive work studying various settings under different priors and inference techniques \citep{grover2018uncertainty, mardani2018neural, heckel2018deep, mixon2018sunlayer, pandit2019asymptotics,lucas2018using,shah2018solving,liu2020information,kabkab2018task,mousavi2018data, raj2019gan,sun2019block}.
In particular, we note that \citet{asim2019invertible} utilize a flow-based prior similar to our setting.

It is important to note that the above approaches focus on recovering a single point estimate that best matches the measurements. However, there can be many inputs that fit the measurements and thus uncertainty in the reconstruction. 
Due to this shortcoming, several recent works focused on recovering the signal distribution conditioned on the measurements \citep{tonolini2019variational, zhang2019probabilistic, adler2018deep, adler2019deep}.

We note that our approach differs from these, since they are learning-based methods that require access to the training data.
On the contrary, our work leverages a \textit{pre-trained} prior to produce an approximate conditional posterior, which can then be used for a variety of tasks such as generating conditional samples or estimating the MMSE recovery.

%-------------------------------------------------------------------------------%
\section{Experiments}

\begin{figure*}[!ht]
\centering
\label{fig:fid_samples_celeba}
\includegraphics[width=0.95\linewidth]{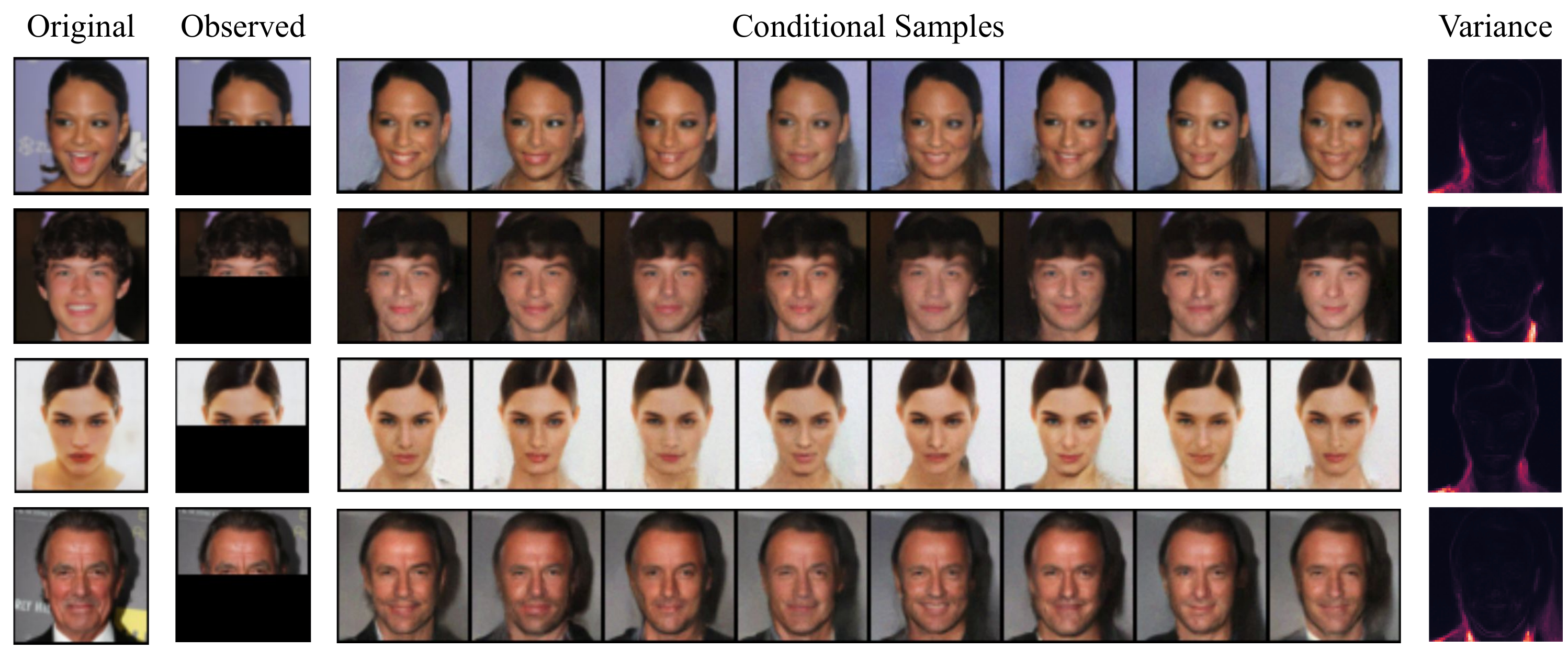}
\caption{Conditional samples generated by our method from observing the upper half of CelebA-HQ faces. We see that our approach is able to produce diverse completions with different jaw lines, mouth shapes, and facial expression. }
\end{figure*}

We validate the efficacy of our proposed method in terms of both sample and reconstruction quality against three baselines: Langevin Monte Carlo (LMC), Ambient VI, and PL-MCMC \citep{cannella2020projected}.  Both LMC and Pl-MCMC are MCMC techniques that can (asymptotically) sample from the true conditional distribution our method tries to approximate. For the comparisons to be fair, we implemented both methods to run MCMC chains in the latent space of the base model, analogous what our method does for VI.  Ambient VI is identical to our method, except it performs VI in the image space and is included for completeness.  In addition, we also conduct our experiments on three different datasets (MNIST, CIFAR-10, and CelebA-HQ) to ensure that our method works across a range of settings.

We report four different sample quality metrics: Frechet Inception Distance (FID), Learned Perceptual Image Patch Similarity (LPIPS), and Inception Score (IS) for CIFAR-10 \citep{heusel2017gans,zhang2018unreasonable,salimans2016improved}. While not strictly a measure of perceptual similarity, the average mean squared error (MSE) is reported for completeness. Additionally, we also report pairwise LPIPS metric used by \citet{zheng2019pluralistic} to measure the diversity of generated samples.

For all our experiments, we use the multiscale RealNVP architecture \citep{dinh2016density} for both the base model and the pre-generator.  We use Adam optimizer \citep{kingma2014adam} to optimize the weights of the pre-generator using the loss defined in \cref{eqn:latent_vi_loss_smooth}. The images used to generate observations were taken from the test set and were not used to train the base models.
We refer the reader to \Cref{appendix:experiment_details} for model hyperparameters and other details of our experiment setup.

\subsection{Image Inpainting}
\label{sec:exp_inpainting}

We perform inpainting tasks using our approach, where we sample missing pixels conditioned on the visible ones. We consider three different conditioning schemes: bottom half (MNIST), top half (CelebA-HQ), and randomly chosen subpixels (CIFAR-10). 
For MNIST, we use the smoothing parameter value of $\sigma=0.1$ and for CIFAR-10 and CelebA-HQ, we use $\sigma=0.05$. 

In \Cref{fig:fid_samples_celeba} we see that our approach produces natural and diverse samples for the missing part of the image. The empirical pixelwise variance (normalized and averaged over the color channels) also confirms that, while the observation is not perfectly matched, most of the high-variance regions are in the unobserved parts as expected.

We also quantitatively evaluate the quality of the generated samples using widely used sample quality metrics, as shown in \Cref{table:fid}. As we can see, our method outperforms the baseline methods on most of the metrics. %For completeness, we include samples for other datasets and methods in \Cref{appendix:extra_results}.
Note that PL-MCMC results for CIFAR-10 and CelebA-HQ are omitted because it was prohibitively slow to run for hundreds of images, as each MCMC chain required over 20,000 proposals. \citet{cannella2020projected} also report using 25,000 proposals for their experiments.  %We define the sample quality metrics used in our experiments in \Cref{appendix:experiment_details}.

Although our method achieves slightly lower diversity value compared to the baselines, we point out that our method also fits the ground truth better (as evidenced by lower MSE and LPIPS to the ground truth). Measuring diversity alone has limitations, as a model could achieve high diversity by inpainting with random noise. 
Thus, we emphasize that in \Cref{fig:cs_psnr}, we observe a noticeable performance gap between using a single sample and the conditional mean (i.e. the optimal MMSE estimator) obtained by averaging multiple samples. This shows high reconstruction accuracy as well as diversity of our samples, since the samples must be \textbf{both diverse and close to the ground truth} for this gap to exist.

\subsection{Compressed Sensing}
\label{sec:exp_cs}

We also present compressed sensing results on the CelebA-HQ dataset. We compare our method to \citep{bora2017compressed} and \citep{asim2019invertible}, two representative techniques for solving inverse problems with a deep generative prior. We did not explicitly compare to the classical sparsity-based priors, as these papers have already demonstrated the superior performance of deep generative priors over them.

Further, we test whether using the conditional mean $\E\brac{\bx\mid\by=\ystar}$ helps or not by evaluating our method in two different ways.  First, we compute the PSNR for individual samples averaged over 32 draws, labeled ``Ours (single)''. Second, we compute the PSNR using the empirical mean of the same 32 samples, labeled ``Ours (MMSE)''.

\begin{figure}[!ht]
\centering
\includegraphics[width=0.95\linewidth]{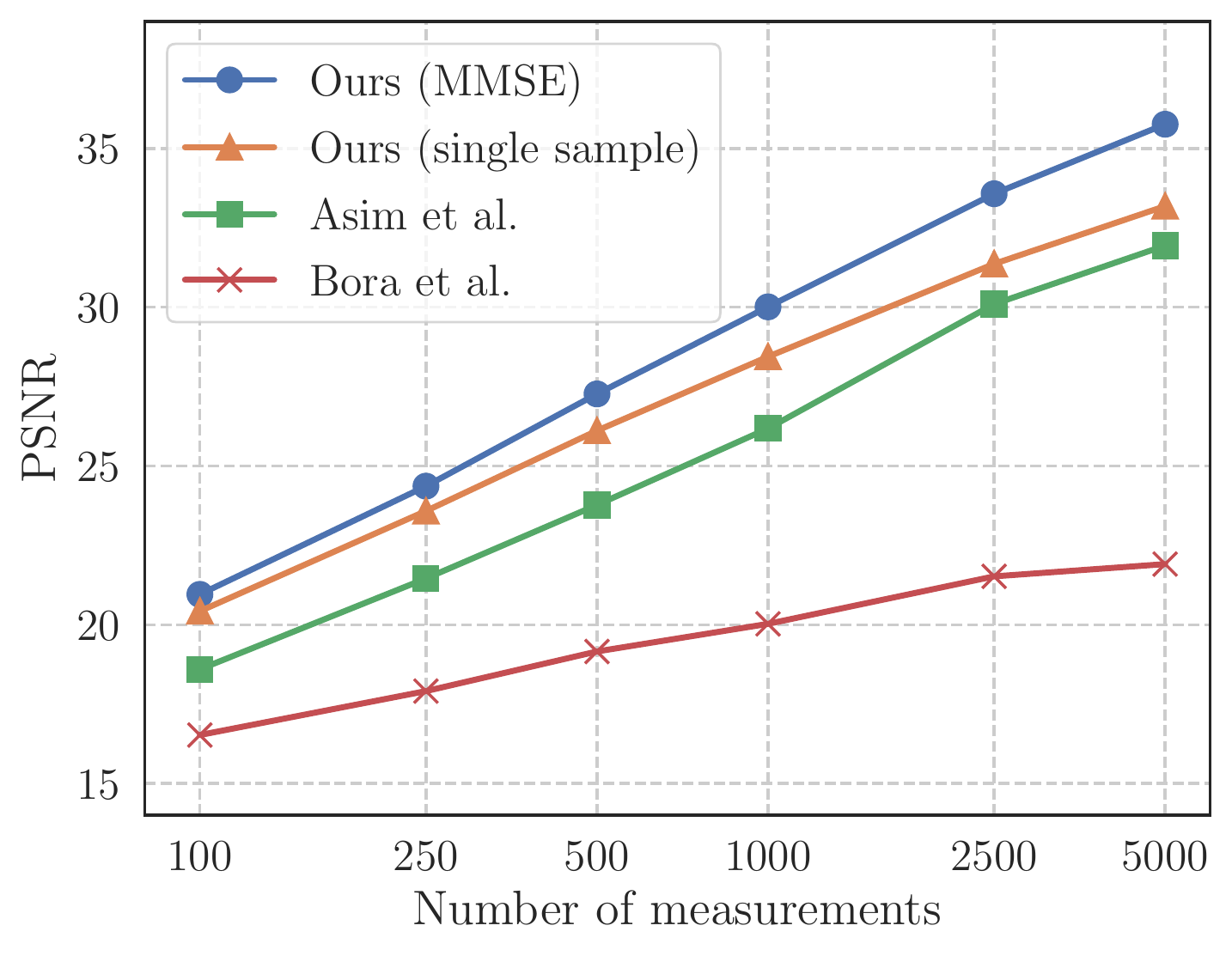}
\caption{Compressed sensing results at varying numbers of measurements.  The plot not only shows that our method outperforms existing methods on a single-sample basis, but also confirms the benefits of using the MMSE estimator.}
\label{fig:cs_psnr}
\end{figure}

As shown in \Cref{fig:cs_psnr}, we notice a significant increase in the PSNR of the recovered signal, especially when using the MMSE estimator.  The relative performance among the presented methods confirms the benefits of \textit{distributional} approaches to inverse problems, and the importance of using the MMSE objective.

\subsection{Uncertainty Quantification}

A key advantage of our approach compared to MCMC-based sampling and existing point estimate methods is the ability to efficiently sample from the learned conditional distribution. This allows us to perform uncertainty quantification by empirically estimating per-pixel variance from a large number of samples.

We demonstrate this in \Cref{fig:pixel_hist}. To create this figure, we took a conditional distribution from the above image completion task and generated 3200 i.i.d. samples. With our flow-based approximate posterior, this only takes 55 seconds \footnote{Measured on a single NVidia GTX 2080 GPU.}. Then we performed kernel density estimation on the histogram of pixel intensity values for each pixel position.

In the figure, we show this result on three representative pixels, each exhibiting a widely different behavior.  As expected, the top pixel in the observed region is sharply concentrated around a single value, where the two unobserved pixels have higher entropy. The pixel at the bottom is a particularly interesting case, as there is some semantic ambiguity given only the top half: it could be part of either the neck or the black background. We see that our learned conditional distribution correctly captures this bimodality, as confirmed by the bottom left plot with highest entropy.

\subsection{Effect of Amortization}

Here we study the effect of amortizing the pre-generator over the observation.  We repeat the image completion experiments from \Cref{sec:exp_inpainting}, except we use a pre-generator that takes in the observed half of the image as conditioning input. The architecture is similar to the conditional flow used in \citep{ardizzone2019guided}, except our model is based on RealNVP instead of Glow.  As a baseline, we consider Ambient VI as well as \citet{asim2019invertible}, also known in the literature as ``inference via optimization'' (labeled IvOM in \Cref{table:avi_fid}) \citep{srivastava2017veegan,metz2017unrolled}.

\begin{figure}[!h]
\centering
\includegraphics[width=\linewidth]{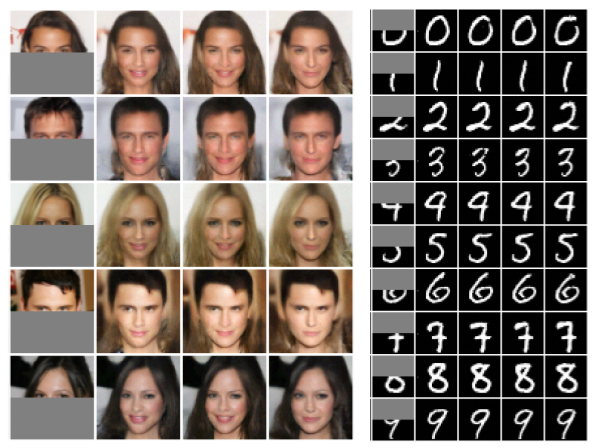}
\caption{Samples generated using the amortized pre-generator.  Each row contains samples conditioned on the masked input in the first column.  We see that the amortized pre-generator produces visually plausible samples without having to perform VI for each observation separately.}
\label{fig:avi_samples}
\end{figure}

The results are shown in \Cref{table:avi_fid} and \Cref{fig:avi_samples}. Compared to the non-amortized version, there is some degradation in sample quality both visually and in terms of FID.  However, the amortized pre-generator significantly improves the inference speed. 

\addtolength{\tabcolsep}{-2pt} 
\begin{table}[!ht]
\centering
\caption{Sample quality and inference speed using the amortized pre-generator. Inference speed is measured as the average time (in seconds) taken to generate a conditional sample.  SVI and LMC results from the image completion experiments in \Cref{sec:exp_inpainting} are provided for comparison.  We can see that amortized inference is several orders of magnitude faster.  We also observe that IvOM \citep{srivastava2017veegan,metz2017unrolled} performs similarly to LMC.}
\vspace{-0.5em}
\label{table:avi_fid}
\begin{tabular}{@{}lcccccc@{}}
\toprule
 & \multicolumn{3}{c}{MNIST} & \multicolumn{3}{c}{CelebA-HQ} \\ 
\cmidrule(r){2-4}
\cmidrule(l){5-7}
& FID & LPIPS & Speed & FID & LPIPS & Speed \\
\midrule
Ours (AVI)  & 8.25 & 0.249 & \textbf{0.025} & 83.1 & 0.463 & \textbf{0.046} \\
Ours (SVI)  & 5.15 & 0.076 & 13.6 & 37.2 & 0.207 & 70.5 \\
LMC         & 14.6 & 0.135 & 3.21 & 30.3 & 0.229 & 88.4 \\
\cmidrule{1-7}
{\small Ambient VI} & 123 & 0.282 & 9.85 & 295 & 0.738 & 67.9 \\
IvOM        & 24.7  & 0.141 & 1.65 & 95.8  & 0.237 & 88.8 \\
% \addlinespace[0.3em]
\bottomrule
\end{tabular}
\end{table}
\vspace{-0.5em}
\addtolength{\tabcolsep}{2pt} 

\begin{figure*}[!ht]
\centering
\includegraphics[width=0.95\linewidth]{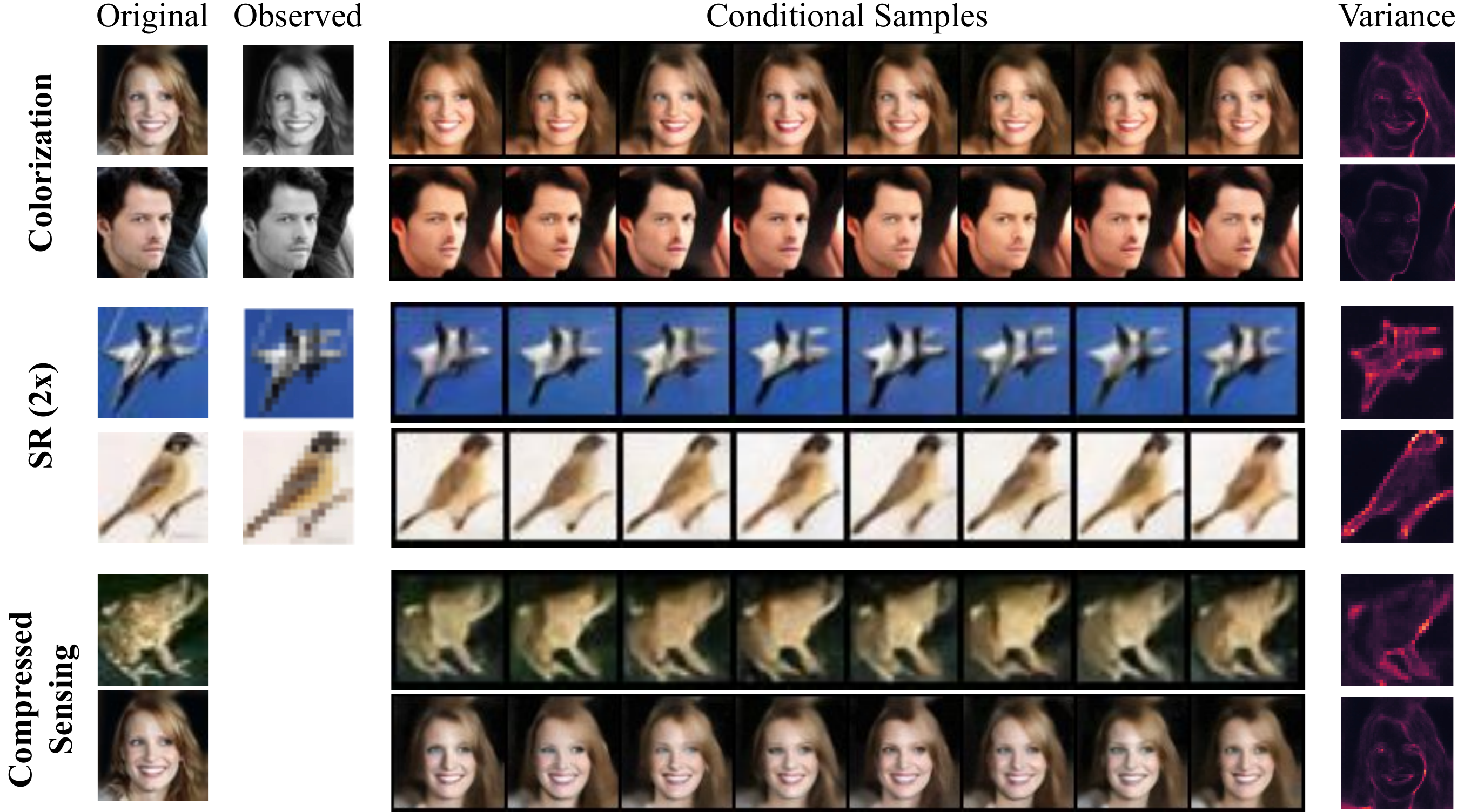}
\caption{Results on various inverse problem tasks using our method.}
\label{fig:inverse_samples}
\end{figure*}

\subsection{Other Inverse Problems}
\label{sec:exp_inverse}
Here we evaluate the versatility of our method on additional linear inverse problems. In \Cref{fig:inverse_samples}, we show the conditional samples obtained by our method on three different tasks: image colorization, super-resolution ($2\times$), and compressed sensing with 500 random Gaussian measurements (for reference, CIFAR-10 images have 3072 dimensions). We notice that the generated samples look natural, even when they do not match the original input perfectly, again showing our method's capability to generate semantically meaningful conditional samples with diversity. 

\subsection{Effects of the Smoothing Parameter}

\begin{figure} \centering
\includegraphics[width=0.90\linewidth]{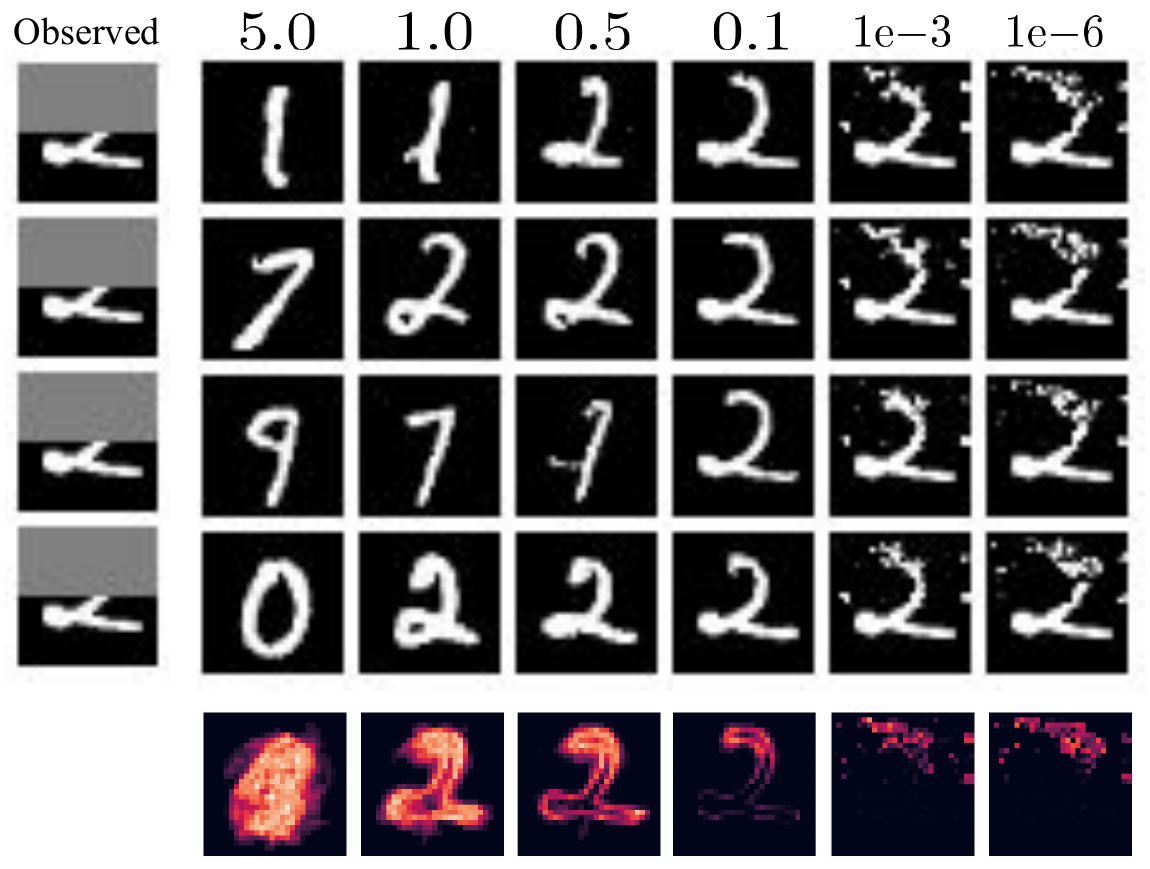}
\caption{Each column contains samples from the learned conditional sampler at different values of $\sigma$ with pixelwise variance computed using 32 samples.}
\label{fig:sigma_effect_samples}
\end{figure}

\begin{figure}
\centering
\includegraphics[width=0.90\linewidth]{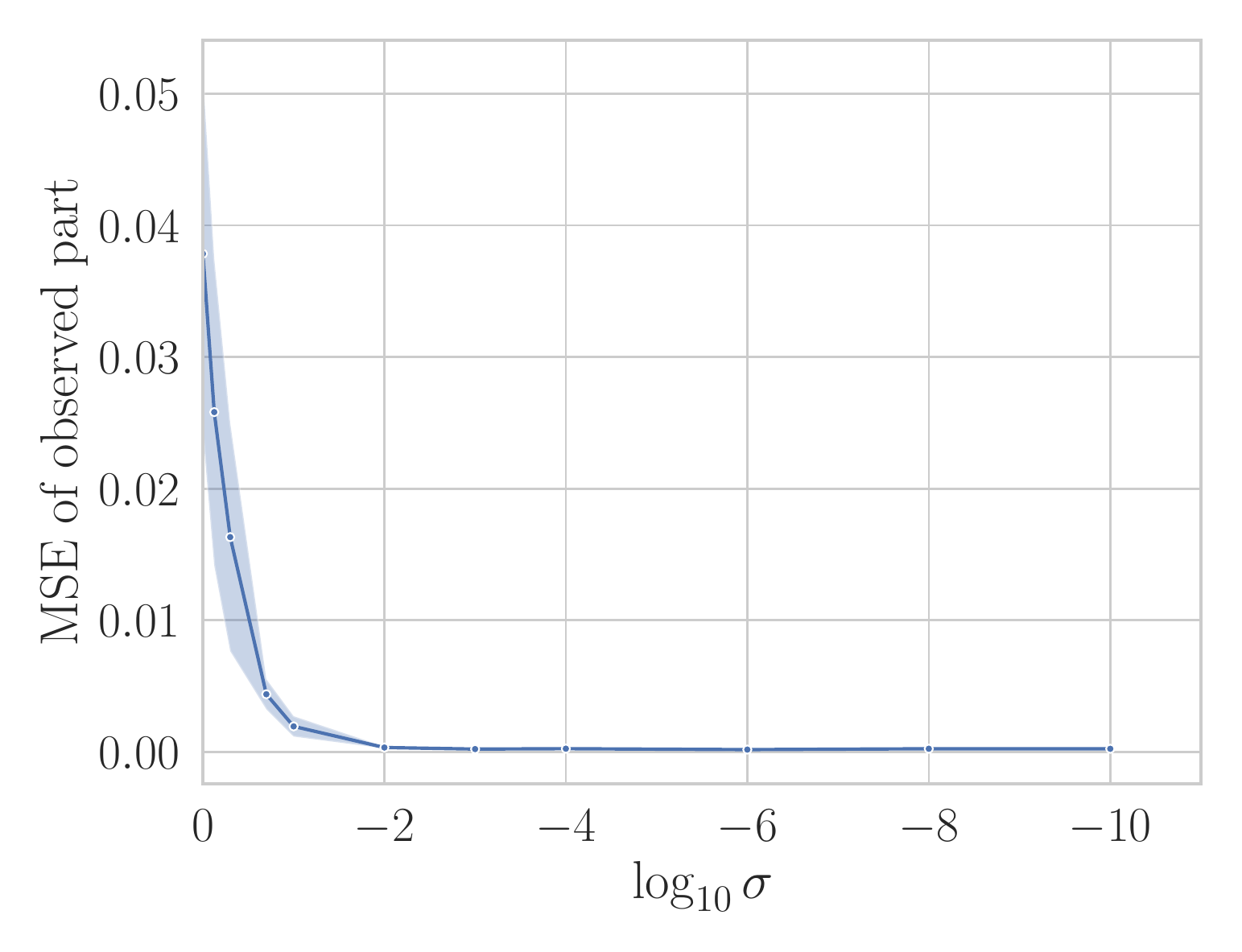}
\caption{MSE between $\bx$ and $\xstar$ at different values of $\sigma$.}
\label{fig:sigma_effect_mse}
\end{figure}

The choice of variance for Gaussian smoothing in \cref{eqn:latent_vi_loss_smooth} is an important hyperparameter, so we provide some empirical analysis on the effects of $\sigma$.  
As shown in \Cref{fig:sigma_effect_samples}, large values of $\sigma$ cause the samples to ignore the observation, whereas small values lead to unnatural samples as the learned distribution tries to match a degenerate true posterior.  Visually, we achieve essentially negligible variance on the observed portion past $\sigma = 0.01$, but at the slight degradation in the sample quality.
In \Cref{fig:sigma_effect_mse}, we also notice that the difference between the true observation ($\xobs^*$) and generated observation ($\xtil$) stops improving past $\sigma = 1\mathrm{e}{-}4$.
We tried annealing $\sigma$ from a large value to a small positive target value to see if that would help improve the sample quality at very small values of $\sigma$, but noticed no appreciable difference.
In practice, we recommend using the largest possible $\sigma$ that produces observations that are within the (task-specific) acceptable range of the true observation.

% \begin{figure} \centering
% \begin{subfigure}{\columnwidth}
%   \centering
%   \includegraphics[width=0.9\linewidth]{images/fig_sigma_effect_samples.pdf}
%   \caption{Each column contains samples from the learned conditional sampler at different values of $\sigma$ with pixelwise variance computed using 32 samples.}
%   \label{fig:sigma_effect_samples}
% \end{subfigure}
% \begin{subfigure}{\columnwidth}
%   \centering
%   \includegraphics[width=0.9\linewidth]{images/fig_sigma_effect_mse.pdf}
%   \caption{MSE between $\xtil$ and $\xobs^*$ at different values of $\sigma$.}
%   \label{fig:sigma_effect_mse}
% \end{subfigure}
% \caption{Effect of the smoothing parameter on sample quality and tightness of approximate conditioning.}
% \label{fig:sigma_effect}
% \end{figure}

%-------------------------------------------------------------------------------%
\section{Conclusion}

We proposed a new technique for solving inverse problems with a normalizing flow prior by viewing them as conditional inference tasks.  The need for approximate inference is motivated by our theoretical hardness result for exact inference.  The particular parametrization of our approximate posterior as a composition of flows is amenable to uncertainty quantification.  We also presented a detailed empirical evaluation of our method with both quantitative and qualitative results on a wide range of tasks and datasets. Further, we show that our formulation can be amortized to significantly improve the inference speed without significantly sacrificing sample quality.
Overall, we believe that the idea of a pre-generator creating structured noise is a useful and general method for solving inverse problems with the benefit of leveraging pre-trained models and quantifying reconstruction uncertainty.

\section*{Acknowledgements}
This research has been supported by NSF Grants CCF 1934932, AF 1901292,
2008710, 2019844 the NSF IFML 2019844 award as well as research gifts by Western Digital, WNCG and MLL, computing resources from TACC and the Archie Straiton Fellowship.

\clearpage
\bibliography{main}

\begin{thebibliography}{68}
\providecommand{\natexlab}[1]{#1}
\providecommand{\url}[1]{\texttt{#1}}
\expandafter\ifx\csname urlstyle\endcsname\relax
  \providecommand{\doi}[1]{doi: #1}\else
  \providecommand{\doi}{doi: \begingroup \urlstyle{rm}\Url}\fi

\bibitem[Adler \& {\"O}ktem(2018)Adler and {\"O}ktem]{adler2018deep}
Adler, J. and {\"O}ktem, O.
\newblock Deep bayesian inversion.
\newblock \emph{arXiv preprint arXiv:1811.05910}, 2018.

\bibitem[Adler \& {\"O}ktem(2019)Adler and {\"O}ktem]{adler2019deep}
Adler, J. and {\"O}ktem, O.
\newblock Deep posterior sampling: Uncertainty quantification for large scale
  inverse problems.
\newblock In \emph{International Conference on Medical Imaging with Deep
  Learning--Extended Abstract Track}, 2019.

\bibitem[Ardizzone et~al.(2019)Ardizzone, L{\"u}th, Kruse, Rother, and
  K{\"o}the]{ardizzone2019guided}
Ardizzone, L., L{\"u}th, C., Kruse, J., Rother, C., and K{\"o}the, U.
\newblock Guided image generation with conditional invertible neural networks.
\newblock \emph{arXiv preprint arXiv:1907.02392}, 2019.

\bibitem[Asim et~al.(2019)Asim, Ahmed, and Hand]{asim2019invertible}
Asim, M., Ahmed, A., and Hand, P.
\newblock Invertible generative models for inverse problems: mitigating
  representation error and dataset bias.
\newblock \emph{arXiv preprint arXiv:1905.11672}, 2019.

\bibitem[Atanov et~al.(2019)Atanov, Volokhova, Ashukha, Sosnovik, and
  Vetrov]{atanov2019semi}
Atanov, A., Volokhova, A., Ashukha, A., Sosnovik, I., and Vetrov, D.
\newblock Semi-conditional normalizing flows for semi-supervised learning.
\newblock \emph{arXiv preprint arXiv:1905.00505}, 2019.

\bibitem[Baraniuk(2007)]{baraniuk2007compressive}
Baraniuk, R.~G.
\newblock Compressive sensing.
\newblock \emph{IEEE signal processing magazine}, 24\penalty0 (4), 2007.

\bibitem[Behrmann et~al.(2019)Behrmann, Duvenaud, and
  Jacobsen]{behrmann2018invertible}
Behrmann, J., Duvenaud, D., and Jacobsen, J.-H.
\newblock Invertible residual networks.
\newblock \emph{International Conference on Machine Learning}, 2019.

\bibitem[Belghazi et~al.(2019)Belghazi, Oquab, LeCun, and
  Lopez-Paz]{belghazi2019learning}
Belghazi, M.~I., Oquab, M., LeCun, Y., and Lopez-Paz, D.
\newblock Learning about an exponential amount of conditional distributions.
\newblock \emph{arXiv preprint arXiv:1902.08401}, 2019.

\bibitem[Bickel et~al.(2009)Bickel, Ritov, Tsybakov,
  et~al.]{bickel2009simultaneous}
Bickel, P.~J., Ritov, Y., Tsybakov, A.~B., et~al.
\newblock Simultaneous analysis of lasso and dantzig selector.
\newblock \emph{The Annals of Statistics}, 37\penalty0 (4):\penalty0
  1705--1732, 2009.

\bibitem[Blei et~al.(2017)Blei, Kucukelbir, and McAuliffe]{blei2017variational}
Blei, D.~M., Kucukelbir, A., and McAuliffe, J.~D.
\newblock Variational inference: A review for statisticians.
\newblock \emph{Journal of the American Statistical Association}, 112\penalty0
  (518):\penalty0 859--877, 2017.

\bibitem[Bora et~al.(2017)Bora, Jalal, Price, and Dimakis]{bora2017compressed}
Bora, A., Jalal, A., Price, E., and Dimakis, A.~G.
\newblock Compressed sensing using generative models.
\newblock In \emph{International Conference on Machine Learning}, pp.\
  537--546. JMLR. org, 2017.

\bibitem[Candes et~al.(2006)Candes, Romberg, and Tao]{candes2006stable}
Candes, E.~J., Romberg, J.~K., and Tao, T.
\newblock Stable signal recovery from incomplete and inaccurate measurements.
\newblock \emph{Communications on Pure and Applied Mathematics: A Journal
  Issued by the Courant Institute of Mathematical Sciences}, 59\penalty0
  (8):\penalty0 1207--1223, 2006.

\bibitem[Cannella et~al.(2020)Cannella, Soltani, and
  Tarokh]{cannella2020projected}
Cannella, C., Soltani, M., and Tarokh, V.
\newblock Projected latent markov chain monte carlo: Conditional sampling of
  normalizing flows, 2020.

\bibitem[Cremer et~al.(2018)Cremer, Li, and Duvenaud]{cremer2018inference}
Cremer, C., Li, X., and Duvenaud, D.
\newblock Inference suboptimality in variational autoencoders.
\newblock In \emph{International Conference on Machine Learning}, pp.\
  1078--1086. PMLR, 2018.

\bibitem[Dinh et~al.(2015)Dinh, Krueger, and Bengio]{dinh2014nice}
Dinh, L., Krueger, D., and Bengio, Y.
\newblock Nice: Non-linear independent components estimation.
\newblock In \emph{International Conference on Learning Representations 2015
  workshop track}, 2015.

\bibitem[Dinh et~al.(2016)Dinh, Sohl-Dickstein, and Bengio]{dinh2016density}
Dinh, L., Sohl-Dickstein, J., and Bengio, S.
\newblock Density estimation using {Real} {NVP}.
\newblock In \emph{International Conference on Learning Representations}, 2016.

\bibitem[Donoho et~al.(2006)]{donoho2006compressed}
Donoho, D.~L. et~al.
\newblock Compressed sensing.
\newblock \emph{IEEE Transactions on information theory}, 52\penalty0
  (4):\penalty0 1289--1306, 2006.

\bibitem[Durkan et~al.(2019)Durkan, Bekasov, Murray, and
  Papamakarios]{durkan2019neural}
Durkan, C., Bekasov, A., Murray, I., and Papamakarios, G.
\newblock Neural spline flows.
\newblock In \emph{Neural Information Processing Systems}, 2019.

\bibitem[Engel et~al.(2017)Engel, Hoffman, and Roberts]{engel2017latent}
Engel, J., Hoffman, M., and Roberts, A.
\newblock Latent constraints: Learning to generate conditionally from
  unconditional generative models.
\newblock \emph{arXiv preprint arXiv:1711.05772}, 2017.

\bibitem[Grover \& Ermon(2019)Grover and Ermon]{grover2018uncertainty}
Grover, A. and Ermon, S.
\newblock Uncertainty autoencoders: Learning compressed representations via
  variational information maximization.
\newblock In \emph{International Conference on Artificial Intelligence and
  Statistics}, 2019.

\bibitem[Heckel \& Hand(2019)Heckel and Hand]{heckel2018deep}
Heckel, R. and Hand, P.
\newblock Deep decoder: {Concise} image representations from untrained
  non-convolutional networks.
\newblock In \emph{International Conference on Learning Representations}, 2019.

\bibitem[Heusel et~al.(2017)Heusel, Ramsauer, Unterthiner, Nessler, and
  Hochreiter]{heusel2017gans}
Heusel, M., Ramsauer, H., Unterthiner, T., Nessler, B., and Hochreiter, S.
\newblock Gans trained by a two time-scale update rule converge to a local nash
  equilibrium.
\newblock In \emph{Advances in neural information processing systems}, pp.\
  6626--6637, 2017.

\bibitem[Ho et~al.(2019)Ho, Chen, Srinivas, Duan, and Abbeel]{ho2019flow++}
Ho, J., Chen, X., Srinivas, A., Duan, Y., and Abbeel, P.
\newblock Flow++: Improving flow-based generative models with variational
  dequantization and architecture design.
\newblock In \emph{International Conference on Machine Learning}, pp.\
  2722--2730, 2019.

\bibitem[Hoffman et~al.(2019)Hoffman, Sountsov, Dillon, Langmore, Tran, and
  Vasudevan]{hoffman2019neutra}
Hoffman, M., Sountsov, P., Dillon, J.~V., Langmore, I., Tran, D., and
  Vasudevan, S.
\newblock Neutra-lizing bad geometry in hamiltonian monte carlo using neural
  transport.
\newblock \emph{arXiv preprint arXiv:1903.03704}, 2019.

\bibitem[Ivanov et~al.(2018)Ivanov, Figurnov, and
  Vetrov]{ivanov2018variational}
Ivanov, O., Figurnov, M., and Vetrov, D.
\newblock Variational autoencoder with arbitrary conditioning.
\newblock \emph{arXiv preprint arXiv:1806.02382}, 2018.

\bibitem[Jordan et~al.(1999)Jordan, Ghahramani, Jaakkola, and
  Saul]{jordan1999introduction}
Jordan, M.~I., Ghahramani, Z., Jaakkola, T.~S., and Saul, L.~K.
\newblock An introduction to variational methods for graphical models.
\newblock \emph{Machine learning}, 37\penalty0 (2):\penalty0 183--233, 1999.

\bibitem[Kabkab et~al.(2018)Kabkab, Samangouei, and Chellappa]{kabkab2018task}
Kabkab, M., Samangouei, P., and Chellappa, R.
\newblock Task-aware compressed sensing with generative adversarial networks.
\newblock In \emph{Proceedings of the AAAI Conference on Artificial
  Intelligence}, volume~32, 2018.

\bibitem[Kingma \& Ba(2014)Kingma and Ba]{kingma2014adam}
Kingma, D.~P. and Ba, J.
\newblock Adam: A method for stochastic optimization.
\newblock \emph{arXiv preprint arXiv:1412.6980}, 2014.

\bibitem[Kingma \& Dhariwal(2018)Kingma and Dhariwal]{kingma2018glow}
Kingma, D.~P. and Dhariwal, P.
\newblock Glow: {Generative} flow with invertible 1x1 convolutions.
\newblock In \emph{Neural Information Processing Systems}, pp.\  10215--10224,
  2018.

\bibitem[Kingma \& Welling(2013)Kingma and Welling]{kingma2013auto}
Kingma, D.~P. and Welling, M.
\newblock Auto-encoding variational bayes.
\newblock \emph{arXiv preprint arXiv:1312.6114}, 2013.

\bibitem[Kingma et~al.(2016)Kingma, Salimans, Jozefowicz, Chen, Sutskever, and
  Welling]{kingma2016improved}
Kingma, D.~P., Salimans, T., Jozefowicz, R., Chen, X., Sutskever, I., and
  Welling, M.
\newblock Improved variational inference with inverse autoregressive flow.
\newblock In \emph{Neural Information Processing Systems}, pp.\  4743--4751,
  2016.

\bibitem[Li et~al.(2019)Li, Akbar, and Oliva]{li2019flow}
Li, Y., Akbar, S., and Oliva, J.~B.
\newblock Flow models for arbitrary conditional likelihoods.
\newblock \emph{arXiv preprint arXiv:1909.06319}, 2019.

\bibitem[Liu \& Scarlett(2020)Liu and Scarlett]{liu2020information}
Liu, Z. and Scarlett, J.
\newblock Information-theoretic lower bounds for compressive sensing with
  generative models.
\newblock \emph{IEEE Journal on Selected Areas in Information Theory},
  1\penalty0 (1):\penalty0 292--303, 2020.

\bibitem[Lucas et~al.(2018)Lucas, Iliadis, Molina, and
  Katsaggelos]{lucas2018using}
Lucas, A., Iliadis, M., Molina, R., and Katsaggelos, A.~K.
\newblock Using deep neural networks for inverse problems in imaging: beyond
  analytical methods.
\newblock \emph{IEEE Signal Processing Magazine}, 35\penalty0 (1):\penalty0
  20--36, 2018.

\bibitem[Mardani et~al.(2018)Mardani, Sun, Donoho, Papyan, Monajemi,
  Vasanawala, and Pauly]{mardani2018neural}
Mardani, M., Sun, Q., Donoho, D., Papyan, V., Monajemi, H., Vasanawala, S., and
  Pauly, J.
\newblock Neural proximal gradient descent for compressive imaging.
\newblock In \emph{Neural Information Processing Systems}, pp.\  9573--9583,
  2018.

\bibitem[Menon et~al.(2020)Menon, Damian, Hu, Ravi, and Rudin]{pulse}
Menon, S., Damian, A., Hu, S., Ravi, N., and Rudin, C.
\newblock Pulse: Self-supervised photo upsampling via latent space exploration
  of generative models.
\newblock \emph{2020 IEEE/CVF Conference on Computer Vision and Pattern
  Recognition (CVPR)}, 2020.
\newblock \doi{10.1109/cvpr42600.2020.00251}.
\newblock URL \url{http://dx.doi.org/10.1109/cvpr42600.2020.00251}.

\bibitem[Metz et~al.(2017)Metz, Poole, Pfau, and
  Sohl{-}Dickstein]{metz2017unrolled}
Metz, L., Poole, B., Pfau, D., and Sohl{-}Dickstein, J.
\newblock Unrolled generative adversarial networks.
\newblock In \emph{5th International Conference on Learning Representations,
  {ICLR} 2017, Toulon, France, April 24-26, 2017, Conference Track
  Proceedings}, 2017.

\bibitem[Mirza \& Osindero(2014)Mirza and Osindero]{mirza2014conditional}
Mirza, M. and Osindero, S.
\newblock Conditional generative adversarial nets.
\newblock \emph{arXiv preprint arXiv:1411.1784}, 2014.

\bibitem[Mixon \& Villar(2018)Mixon and Villar]{mixon2018sunlayer}
Mixon, D.~G. and Villar, S.
\newblock Sunlayer: {Stable} denoising with generative networks.
\newblock \emph{arXiv preprint arXiv:1803.09319}, 2018.

\bibitem[Mousavi et~al.(2018)Mousavi, Dasarathy, and Baraniuk]{mousavi2018data}
Mousavi, A., Dasarathy, G., and Baraniuk, R.~G.
\newblock A data-driven and distributed approach to sparse signal
  representation and recovery.
\newblock In \emph{International Conference on Learning Representations}, 2018.

\bibitem[Neal et~al.(2011)]{neal2011mcmc}
Neal, R.~M. et~al.
\newblock Mcmc using hamiltonian dynamics.
\newblock \emph{Handbook of markov chain monte carlo}, 2\penalty0
  (11):\penalty0 2, 2011.

\bibitem[Nijkamp et~al.(2020)Nijkamp, Gao, Sountsov, Vasudevan, Pang, Zhu, and
  Wu]{nijkamp2020learning}
Nijkamp, E., Gao, R., Sountsov, P., Vasudevan, S., Pang, B., Zhu, S.-C., and
  Wu, Y.~N.
\newblock Learning energy-based model with flow-based backbone by neural
  transport mcmc.
\newblock \emph{arXiv preprint arXiv:2006.06897}, 2020.

\bibitem[Ongie et~al.(2020)Ongie, Jalal, Metzler, Baraniuk, Dimakis, and
  Willett]{ongie2020deep}
Ongie, G., Jalal, A., Metzler, C.~A., Baraniuk, R.~G., Dimakis, A.~G., and
  Willett, R.
\newblock Deep learning techniques for inverse problems in imaging.
\newblock \emph{IEEE Journal on Selected Areas in Information Theory},
  1\penalty0 (1):\penalty0 39--56, 2020.

\bibitem[Oord et~al.(2018)Oord, Li, Babuschkin, Simonyan, Vinyals, Kavukcuoglu,
  Driessche, Lockhart, Cobo, Stimberg, et~al.]{oord2018parallel}
Oord, A., Li, Y., Babuschkin, I., Simonyan, K., Vinyals, O., Kavukcuoglu, K.,
  Driessche, G., Lockhart, E., Cobo, L., Stimberg, F., et~al.
\newblock Parallel wavenet: Fast high-fidelity speech synthesis.
\newblock In \emph{International conference on machine learning}, pp.\
  3918--3926. PMLR, 2018.

\bibitem[Pajot et~al.(2019)Pajot, de~Bezenac, and
  Gallinari]{pajot2018unsupervised}
Pajot, A., de~Bezenac, E., and Gallinari, P.
\newblock Unsupervised adversarial image reconstruction.
\newblock In \emph{International Conference on Learning Representations}, 2019.
\newblock URL \url{https://openreview.net/forum?id=BJg4Z3RqF7}.

\bibitem[Pandit et~al.(2019)Pandit, Sahraee, Rangan, and
  Fletcher]{pandit2019asymptotics}
Pandit, P., Sahraee, M., Rangan, S., and Fletcher, A.~K.
\newblock Asymptotics of map inference in deep networks.
\newblock \emph{arXiv preprint arXiv:1903.01293}, 2019.

\bibitem[Papamakarios et~al.(2019)Papamakarios, Nalisnick, Rezende, Mohamed,
  and Lakshminarayanan]{papamakarios2019survey}
Papamakarios, G., Nalisnick, E., Rezende, D.~J., Mohamed, S., and
  Lakshminarayanan, B.
\newblock Normalizing flows for probabilistic modeling and inference.
\newblock \emph{arXiv preprint arXiv:1912.02762}, 2019.

\bibitem[Papamarkou et~al.(2019)Papamarkou, Hinkle, Young, and
  Womble]{papamarkou2019challenges}
Papamarkou, T., Hinkle, J., Young, M.~T., and Womble, D.
\newblock Challenges in bayesian inference via markov chain monte carlo for
  neural networks.
\newblock \emph{arXiv}, pp.\  arXiv--1910, 2019.

\bibitem[Pathak et~al.(2016)Pathak, Krahenbuhl, Donahue, Darrell, and
  Efros]{pathak2016context}
Pathak, D., Krahenbuhl, P., Donahue, J., Darrell, T., and Efros, A.~A.
\newblock Context encoders: Feature learning by inpainting.
\newblock In \emph{Proceedings of the IEEE conference on computer vision and
  pattern recognition}, pp.\  2536--2544, 2016.

\bibitem[Raj et~al.(2019)Raj, Li, and Bresler]{raj2019gan}
Raj, A., Li, Y., and Bresler, Y.
\newblock Gan-based projector for faster recovery with convergence guarantees
  in linear inverse problems.
\newblock In \emph{Proceedings of the IEEE/CVF International Conference on
  Computer Vision}, pp.\  5602--5611, 2019.

\bibitem[Rezende \& Mohamed(2015)Rezende and Mohamed]{rezende2015variational}
Rezende, D. and Mohamed, S.
\newblock Variational inference with normalizing flows.
\newblock In \emph{International Conference on Machine Learning}, pp.\
  1530--1538, 2015.

\bibitem[Richardson et~al.(2020)Richardson, Alaluf, Patashnik, Nitzan, Azar,
  Shapiro, and Cohen-Or]{richardson2020encoding}
Richardson, E., Alaluf, Y., Patashnik, O., Nitzan, Y., Azar, Y., Shapiro, S.,
  and Cohen-Or, D.
\newblock Encoding in style: a stylegan encoder for image-to-image translation.
\newblock \emph{arXiv preprint arXiv:2008.00951}, 2020.

\bibitem[Salimans et~al.(2016)Salimans, Goodfellow, Zaremba, Cheung, Radford,
  and Chen]{salimans2016improved}
Salimans, T., Goodfellow, I., Zaremba, W., Cheung, V., Radford, A., and Chen,
  X.
\newblock Improved techniques for training gans.
\newblock In \emph{Advances in neural information processing systems}, pp.\
  2234--2242, 2016.

\bibitem[Shah \& Hegde(2018)Shah and Hegde]{shah2018solving}
Shah, V. and Hegde, C.
\newblock Solving linear inverse problems using gan priors: An algorithm with
  provable guarantees.
\newblock In \emph{2018 IEEE international conference on acoustics, speech and
  signal processing (ICASSP)}, pp.\  4609--4613. IEEE, 2018.

\bibitem[Sohn et~al.(2015)Sohn, Lee, and Yan]{sohn2015learning}
Sohn, K., Lee, H., and Yan, X.
\newblock Learning structured output representation using deep conditional
  generative models.
\newblock In \emph{Neural Information Processing Systems}, pp.\  3483--3491,
  2015.

\bibitem[Song \& Ermon(2019)Song and Ermon]{song2019generative}
Song, Y. and Ermon, S.
\newblock Generative modeling by estimating gradients of the data distribution.
\newblock In \emph{Advances in Neural Information Processing Systems}, pp.\
  11918--11930, 2019.

\bibitem[Srivastava et~al.(2017)Srivastava, Valkov, Russell, Gutmann, and
  Sutton]{srivastava2017veegan}
Srivastava, A., Valkov, L., Russell, C., Gutmann, M.~U., and Sutton, C.
\newblock Veegan: reducing mode collapse in gans using implicit variational
  learning.
\newblock In \emph{Proceedings of the 31st International Conference on Neural
  Information Processing Systems}, pp.\  3310--3320, 2017.

\bibitem[Sun et~al.(2019)Sun, Liu, and Kamilov]{sun2019block}
Sun, Y., Liu, J., and Kamilov, U.~S.
\newblock Block coordinate regularization by denoising.
\newblock \emph{NeurIPS}, 2019.

\bibitem[Tibshirani(1996)]{tibshirani1996regression}
Tibshirani, R.
\newblock Regression shrinkage and selection via the lasso.
\newblock \emph{Journal of the Royal Statistical Society: Series B
  (Methodological)}, 58\penalty0 (1):\penalty0 267--288, 1996.

\bibitem[Tonolini et~al.(2019)Tonolini, Lyons, Caramazza, Faccio, and
  Murray-Smith]{tonolini2019variational}
Tonolini, F., Lyons, A., Caramazza, P., Faccio, D., and Murray-Smith, R.
\newblock Variational inference for computational imaging inverse problems.
\newblock \emph{arXiv preprint arXiv:1904.06264}, 2019.

\bibitem[Wainwright et~al.(2008)Wainwright, Jordan,
  et~al.]{wainwright2008graphical}
Wainwright, M.~J., Jordan, M.~I., et~al.
\newblock Graphical models, exponential families, and variational inference.
\newblock \emph{Foundations and Trends{\textregistered} in Machine Learning},
  1\penalty0 (1--2):\penalty0 1--305, 2008.

\bibitem[Ward et~al.(2019)Ward, Smofsky, and Bose]{ward2019improving}
Ward, P.~N., Smofsky, A., and Bose, A.~J.
\newblock Improving exploration in soft-actor-critic with normalizing flows
  policies.
\newblock \emph{arXiv preprint arXiv:1906.02771}, 2019.

\bibitem[Welling \& Teh(2011)Welling and Teh]{welling2011bayesian}
Welling, M. and Teh, Y.~W.
\newblock Bayesian learning via stochastic gradient langevin dynamics.
\newblock In \emph{Proceedings of the 28th international conference on machine
  learning (ICML-11)}, pp.\  681--688, 2011.

\bibitem[Whang et~al.(2020)Whang, Lei, and Dimakis]{whang2020compressed}
Whang, J., Lei, Q., and Dimakis, A.~G.
\newblock Compressed sensing with invertible generative models and dependent
  noise.
\newblock \emph{arXiv preprint arXiv:2003.08089}, 2020.

\bibitem[Yu et~al.(2018)Yu, Lin, Yang, Shen, Lu, and Huang]{yu2018generative}
Yu, J., Lin, Z., Yang, J., Shen, X., Lu, X., and Huang, T.~S.
\newblock Generative image inpainting with contextual attention.
\newblock In \emph{Proceedings of the IEEE conference on computer vision and
  pattern recognition}, pp.\  5505--5514, 2018.

\bibitem[Zhang \& Jin(2019)Zhang and Jin]{zhang2019probabilistic}
Zhang, C. and Jin, B.
\newblock Probabilistic residual learning for aleatoric uncertainty in image
  restoration.
\newblock \emph{arXiv preprint arXiv:1908.01010}, 2019.

\bibitem[Zhang et~al.(2018)Zhang, Isola, Efros, Shechtman, and
  Wang]{zhang2018unreasonable}
Zhang, R., Isola, P., Efros, A.~A., Shechtman, E., and Wang, O.
\newblock The unreasonable effectiveness of deep features as a perceptual
  metric.
\newblock In \emph{Proceedings of the IEEE conference on computer vision and
  pattern recognition}, pp.\  586--595, 2018.

\bibitem[Zheng et~al.(2019)Zheng, Cham, and Cai]{zheng2019pluralistic}
Zheng, C., Cham, T.-J., and Cai, J.
\newblock Pluralistic image completion.
\newblock In \emph{Proceedings of the IEEE/CVF Conference on Computer Vision
  and Pattern Recognition}, pp.\  1438--1447, 2019.

\end{thebibliography}
\bibliographystyle{icml2021}

\clearpage
\appendix
\section{Proof of Hardness Results}
\label{appendix:hardness_results}

\subsection{Preliminaries}

A Boolean variable is a variable that takes a value in $\{-1, 1\}$. A \textit{literal} is a Boolean variable $x_i$ or its negation $(\lnot x_i)$. A \textit{clause} is set of literals combined with the OR operator, e.g., $(x_1 \lor \lnot x_2 \lor x_3)$. A \textit{conjunctive normal form formula} is a set of clauses joined by the AND operator, e.g. $(x_1 \lor \lnot x_2 \lor x_3) \land (x_1 \lor \lnot x_3 \lor x_4)$. A satisfying assignment is an assignment to the variables such that the Boolean formula is true.

The \textit{3-SAT problem} is the problem of deciding if a conjunctive normal form formula with three literals per clause has a satisfying assignment. We will show that conditional sampling from flow models allows us to solve the 3-SAT problem.

We ignore the issue of representing samples from the conditional distribution with a finite number of bits. However the reduction is still valid if the samples are truncated to a constant number of bits.

\subsection{Design of the Additive Coupling Network}

Given a conjunctive normal form with $m$ clauses, we design a ReLU neural network with 3 hidden layers such that the output is $0$ if the input is far from a satisfying assignment, and the output is about a large number $M$ if the input is close to a satisfying assignment.

\begin{figure}[ht]
    \centering
    \includegraphics[width=\linewidth]{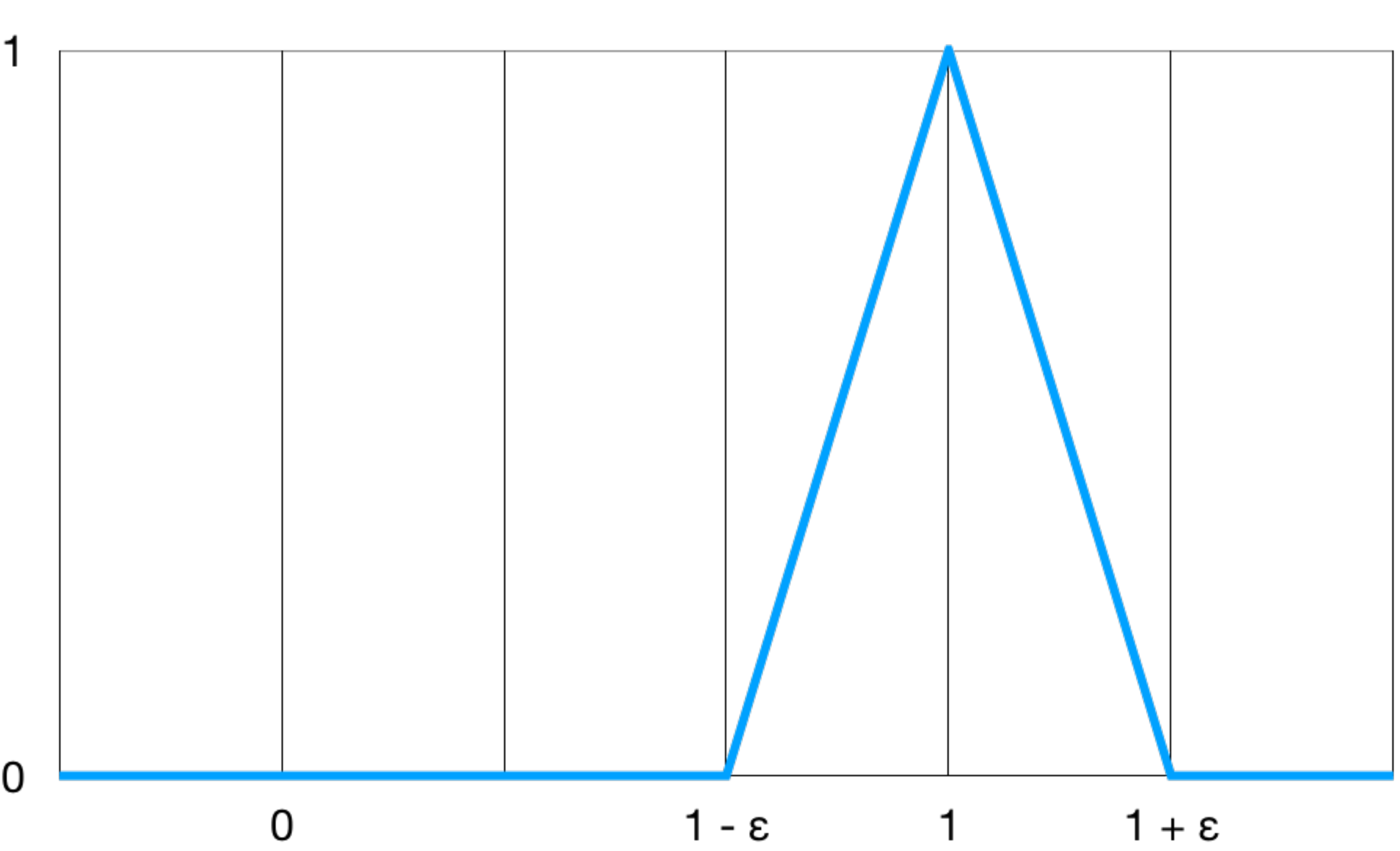}
    \caption{Plot of the scalar function used to construct an additive coupling layer that can generate samples of satisfying 3-SAT assignments.}
    \label{fig:smooth_delta}
\end{figure}

We will define the following scalar function
\begin{align*}
\delta_\varepsilon(x) &= \mathrm{ReLU}\left(\frac{1}{\varepsilon}(x - (1 - \varepsilon))\right) \\
&-\mathrm{ReLU}\left(\frac{1}{\varepsilon}(x - (1 - \varepsilon)) - 1\right) \\
&-\mathrm{ReLU}\left(\frac{1}{\varepsilon}(x - 1)\right) \\
&+\mathrm{ReLU}\left(\frac{1}{\varepsilon}(x - 1) - 1\right).
\end{align*}
This function is $1$ if the input is $1$, $0$ if the input $x$ has $\vert x - 1 \vert \geq \varepsilon$ and is a linear interpolation on $(1 - \varepsilon, 1 + \varepsilon)$. Note that it can be implemented by a hidden layer of a neural network and a linear transform, which can be absorbed in the following hidden layer. See \Cref{fig:smooth_delta} for a plot of this function.

For each variable $x_i$, we create a transformed variable $\tilde{x}_i$ by applying $\tilde{x}_i = \delta_\varepsilon(x_i) - \delta_\varepsilon(-x_i)$. Note that this function is $0$ on $(-\infty, -1 - \varepsilon] \cup [-1 + \varepsilon, 1 - \varepsilon] \cup [1 + \varepsilon, \infty)$, $-1$ at $x_i = -1$, $1$ at $x_i = 1$, and a smooth interpolation on the remaining values in the domain.

Every clause has at most $8$ satisfying assignments. For each satisfying assignment we will create a neuron with the following process: (1) get the relevant transformed values $\tilde{x}_i, \tilde{x}_j, \tilde{x}_k$, (2) multiply each variable by $1/3$ if it is equal to $1$ in the satisfying assignment and $-1/3$ if it is equal to $-1$ in the satisfying assignment, (3) sum the scaled variables, (4) apply the $\delta_\varepsilon$ function to the sum.

We will then sum all the neurons corresponding to a satisfying assignment for clause $C_j$ to get the value $c_j$.
The final output is the value $M \times \mathrm{ReLU}(\sum_j c_j - (m - 1))$, where $M$ is a large scalar.

We say that an input to the neural network $x$ corresponds to a Boolean assignment $x' \in \{-1, 1\}^d$ if for every $x_i$ we have $\vert x_i - x'_i\vert < \varepsilon$. For $\varepsilon < 1/3$, if the input does not correspond to a satisfying assignment of the given formula, then at least one of the values $c_j$ is $0$. The remaining values of $c_j$ are at most $1$, so the sum in the output is at most $(m - 1)$, thus the sum is at most zero, so the final output is $0$. However, if the input is a satisfying assignment, then every value of $c_j = 1$, so the output is $M$.

\subsection{Generating SAT Solutions from the Conditional Distribution}

Our flow model will take in Gaussian noise $x_1, \ldots, x_d, z \sim N(0, 1)$. The values $x_1, \ldots, x_d$ will be passed through to the output. The output variable $y$ will be $z + f_M(x_1, \ldots, x_d)$, where $f_M$ is the neural network described in the previous section, and $M$ is the parameter in the output to be decided later.

Let $A$ be all the valid satisfying assignments to the given formula. For each assignment $a$, we will define $X_a$ to be the region $X_a = \{x \in \mathbb{R}^d : \|a - x\|_\infty \leq \varepsilon\}$, where as above $\varepsilon$ is some constant less than $1/3$. Let $X_A = \bigcup_{a \in A} X_a$.

Given an element $x \in X_a$, we can recreate the corresponding satisfying assignment $a$. Thus if we have an element of $X_A$, we can certify that there is a satisfying assignment. We will show that the distribution conditioned on $y = M$ can generate satisfying assignments with high probability.

We have that
\[
p(X_A \mid y = M) = \frac{p(y = M, X_A)}{p(y = M, X_A) + p(y = M, \setcomp{X}_A)}
\]
If we can show that $p(y = M, \setcomp{X}_A)\ll p(y = M, X_A)$, then we have that the generated samples are with high probability satisfying assignments.

Note that,
\begin{align*}
p(y = M, \setcomp{X}_A)
&= p(y = M \mid \setcomp{X}_A)P(\setcomp{X}_A) \\
&\leq p(y = M \mid \setcomp{X}_A).
\end{align*}
Also notice that if $x \in \setcomp{X}_A$, then $f_M(x) = 0$. Thus $y \sim \gN(0, 1)$ and $P(y = M \mid \setcomp{X}_A) = \Theta(\exp(-M^2/2))$.

Now consider any satisfying assignment $x_a$. Let $X'_a$ be the region $X'_a = \{x \in \mathbb{R}^d : \|a - x\|_\infty \leq \frac{1}{2m}\}$. Note that for every $x$ in this region we have $f_M(x) \geq M/2$. Additionally, we have that $P(X'_a) = \Theta(m)^{-d}$. Thus for any $x \in X'_a$, we have $p(Y = M \mid x) \gtrsim \exp(-M^2/8)$. We can conclude that
\begin{align*}
p(y = M, X_A)
&\geq p(Y = M, X'_a) \\
&= \int_{X'_a}p(Y = M \mid x) p(x)\,dx \\
&\gtrsim \exp(-M^2/8 - \Theta(d \log m)).
\end{align*}
For $M = O(\sqrt{d \log m})$, we have that $p(y = M, \setcomp{X}_A)$ is exponentially smaller than $p(y = M, X_A)$. This implies that sampling from the distribution conditioned on $y = M$ will return a satisfying assignment with high probability.

\subsection{Hardness of Approximate Sampling}

\begin{definition}
The complexity class $RP$ is the class of decision problems with efficient random algorithms that (1) output YES with probability $1/2$ if the true answer is YES and (2) output NO with probability $1$ if the true answer is NO. It is widely believed that $RP$ is a strict subset of $NP$.
\end{definition}

A simple extension of the above theorem shows that even approximately matching the true conditional distribution in terms of the total variation (TV) distance is computationally hard. TV distance is defined as $d_\mathrm{TV}(p, q) = \sup_{E} \vert p(E) - q(E) \vert \le 1$, where $E$ is an event. The below corollary shows that it is hard to conditionally sample from a distribution that is even slightly bounded away from 1.

\begin{corollary}
\label{cor:approximate_hardness}
The conditional sampling problem remains hard even if we only require the algorithm to sample from a distribution $q$ such that $d_\mathrm{TV}(p(\cdot \mid x=x^*), q) \leq 1 -  1/\mathrm{poly}(d)$, where $d$ is the dimension of the distribution.
\end{corollary}

We show that the problem is still hard even if we require the algorithm to sample from a distribution $q$ such that $d_\mathrm{TV}(p(x \mid y=y^*), q) \geq 1/\mathrm{poly}(d)$.

Consider the event $X_A$ from above. We saw that $p(X_A \mid y = M) \geq 1 - \exp(-\Omega(d))$. We have that $d_\mathrm{TV}(p(\cdot \mid y = M), q) \geq 1 - \exp(-\Omega(d) - q(X_A))$.

Suppose that the distribution $q$ has $q(X_A) \geq 1 / \mathrm{poly}(d)$. Then by sampling a polynomial number of times from $q$ we sample an element of $X_A$, which allows us to find a satisfying assignment. Thus if we can efficiently create such a distribution, we would be able to efficiently solve SAT and RP = NP. As we are assuming this is false, we must have $q(X_A) \leq 1 / \mathrm{poly}(d)$, which implies $d_\mathrm{TV}(p(\cdot \mid y = M), q) \geq 1 - 1/\mathrm{poly}(d)$.

\section{Missing Derivations}

\subsection{Derivation of \Cref{eqn:latent_vi_loss_smooth}}
\label{appendix:loss_simplification}

Here we present a detailed derivation of \Cref{eqn:latent_vi_loss_smooth}. Note that this equality is true up to a constant w.r.t. $\fcond$.
\begin{align*}
&\Lours(\fcond) \\
&\triangleq \kldiv{\pcomp(\bx)}{\pbase(\bx\mid\ytil=\ystar)} \\
&= \E_{\bx\sim\pcomp}\brac{\log \pcomp(\bx) - \log \pbase(\bx,\ytil=\ystar)} + \log \pbase(\ytil=\ystar) \\
&\stackeq{A} \E_{\bx\sim\pcomp}\brac{\log \pcomp(\bx) - \log \pbase(\bx) - \log\psig(\ytil=\ystar\mid\bx)} \\
&\stackeq{B} \E_{\bx\sim\pcomp}\brac{\log \pcomp(\bx) - \log \pbase(\bx)} \\
&\hspace{1.5em}+ \E_{\bx\sim\pcomp}\brac{-\log\psig(\ytil=\ystar\mid \by=A(\bx))} \\
&= \kldiv{\pcomp(\bx)}{\pbase(\bx)} \\
&\hspace{1.5em}+ \E_{\bx\sim\pcomp}\brac{-\log\psig(\ytil=\ystar\mid \by=A(\bx))} \\
&\stackeq{C} \kldiv{\pcond(\bz)}{\pbase(\bz)} + \E_{\bz\sim\pcond}\brac{\frac{1}{2\sigma^2}\norm{A(\fbase(\bz))-\ystar}_2^2} 
\end{align*}
In ($A$), we drop $\log \pbase(\ytil=\ystar)$, as it is constant w.r.t. $\fcond$.\\
In ($B$), we use the conditional independence $\ytil \perp\!\!\!\perp \bx \mid \by$.\\
In ($C$), we use the invariance of KL divergence under invertible transformation to rewrite it in terms of $\bz$.

\subsection{Joint VI vs. Marginal VI}
\label{appendix:loss_bound}

We also provide a justification for using the joint VI loss as discussed in \Cref{sec:our_method}. Specifically, we show that the joint VI loss in \cref{eqn:latent_vi_loss_smooth} is an upper bound to the intractable marginal VI loss. Assuming the partitioning $\bx = (\bx_1, \bx_2)$, we have: 
\begin{align*}
&\text{(Joint KL)} \\
&= \kldiv{\pcomp(\bx)}{\pbase(\bx|\xtil=\xstar)} \\
&= \E_{\pcomp} \brac{\log \pcomp(\xobs,\xmis) - \log \pbase (\xobs,\xmis |\xtil = \xstar)} \\
&= \E_{\pcomp} \biggl[\log \pcomp(\xmis) + \log \pcomp(\xobs\mid\xmis) \\
&\hspace{1em}- \log \pbase (\xmis |\xtil = \xstar) - \log \pbase(\xobs\mid\xtil=\xstar,\xmis) \biggr] \\
&= \E_{\pcomp} \brac{\log \pcomp(\xmis) - \log \pbase (\xmis |\xtil = \xstar)} \\
&+ \E_{\pcomp}\biggl[\E_{\pcomp(\xobs\mid\xmis)} \biggl[ \\
&\hspace{2em}\log \pcomp(\xobs\mid\xmis) - \log \pbase(\xobs\mid\xtil=\xstar,\xmis)\biggr]\biggr] \\
&= \kldiv{\pcomp(\xmis)}{\pbase (\xmis|\xtil=\xstar)} \\
&\hspace{1em}+ \mathbb{E}_{\pcomp(\xmis)} \brac{ \kldiv{\pcomp(\xobs \mid \xmis)}{\pbase (\xobs|\xtil=\xstar, \xmis)}} \\
&\ge \kldiv{\pcomp(\xmis)}{\pbase (\xmis|\xtil=\xstar)} \\
&= \text{(Marginal KL)},
\end{align*}
where the last inequality is due to the nonnegativity of KL.
Note that equality holds when
$$\kldiv{\pcomp(\xobs \mid \xmis)}{\pbase (\xobs|\xtil=\xstar, \xmis)} = 0,$$
i.e. when our variational posterior matches the true conditional.

\section{Experiment Details}
\label{appendix:experiment_details}

\subsection{Our Algorithm}

\begin{algorithm}[!h]
	\caption{\small\textbf{Training the pre-generator for a given observation under transformation}. We assume that $\fcond$ is an invertible neural network with parameters $\theta$.}
	\label{alg:main}
	\begin{algorithmic}[1]
        \STATE \textbf{Input}: $\ystar$: observation, $A$: differentiable measurement function.
        \FOR{$i = 1 \ldots \mathtt{num\_steps}$}
            \FOR{$j = 1 \ldots m$}
                \STATE Sample $\eps^{(j)} \sim \Normal$
                \STATE $\bz^{(j)} \gets \fcond(\eps^{(j)})$ \hfill (reparametrization trick)
            \ENDFOR
            \STATE
            $\gL \gets \frac{1}{m}\sum\limits_{j=1}^m \biggl[\log\pcond(\bz^{(j)}) - \log\pnormal(\bz^{(j)})$ \newline
            \hspace*{2cm} $+\frac{1}{2\sigma^2}\norm{A(\fbase(\bz^{(j)}))-\ystar}_2^2 \biggr]$
            \STATE $\theta \gets \theta - \nabla_\theta \gL$ \hfill (gradient step)
        \ENDFOR
        % \For{$i = 1\ldots K$}
        %     \State $x \sim \D$
        %     \State $\lambda_0 \gets f_{\phi_t}(x)$
        %     %\State $\hat{\theta} \gets \lceil \theta \rceil$
        %         \For{$i = 0\ldots k$}
        %             \State $z_i \sim q(z \scolon \lambda_i)$ \Comment{reparameterize as $z_{\lambda_i}(\eps)$}
        %             \State $w(z \scolon \lambda_i, \theta) \gets {p_{{\theta}}(x, z_i)}/{q(z_i \scolon \lambda_i)}$
        %             \If{$i < k$}
        %                 \State $\lambda_{i + 1} \gets \lceil\lambda_i + \eta \nabla_{\lambda_i} \ln w(z \scolon \lambda_i, \theta)\rceil$
        %                 %\State $\lambda_{i + 1} \gets \lambda_i + \eta \nabla_{\lambda_i} \ln w_i$
        %             \EndIf
        %         \EndFor
        %     \State $\phi_{t+1} \gets \phi_t + \nabla_{\phi_t} \ln w(z_0 \scolon \lambda_0, \theta_t)$
        %     \If{Train with SVI}
        %         \State $\theta_{t+1} \gets \theta_t + \nabla_{\theta_t} \ln w(z_k \scolon \lambda_k, \theta_t)$
        %     \ElsIf{Train with BSVI}
        %         \State $\theta_{t+1} \gets \theta_t + \nabla_{\theta_t} \ln \sum_i \pi_i w(z_i \scolon \lambda_i, \theta_t)$
        %     \EndIf
        % \EndFor
	\end{algorithmic}
\end{algorithm}

\subsection{Hyperparameters: Base Model and Pre-generator}
\begin{figure*}[!ht]
\centering
\includegraphics[width=\linewidth]{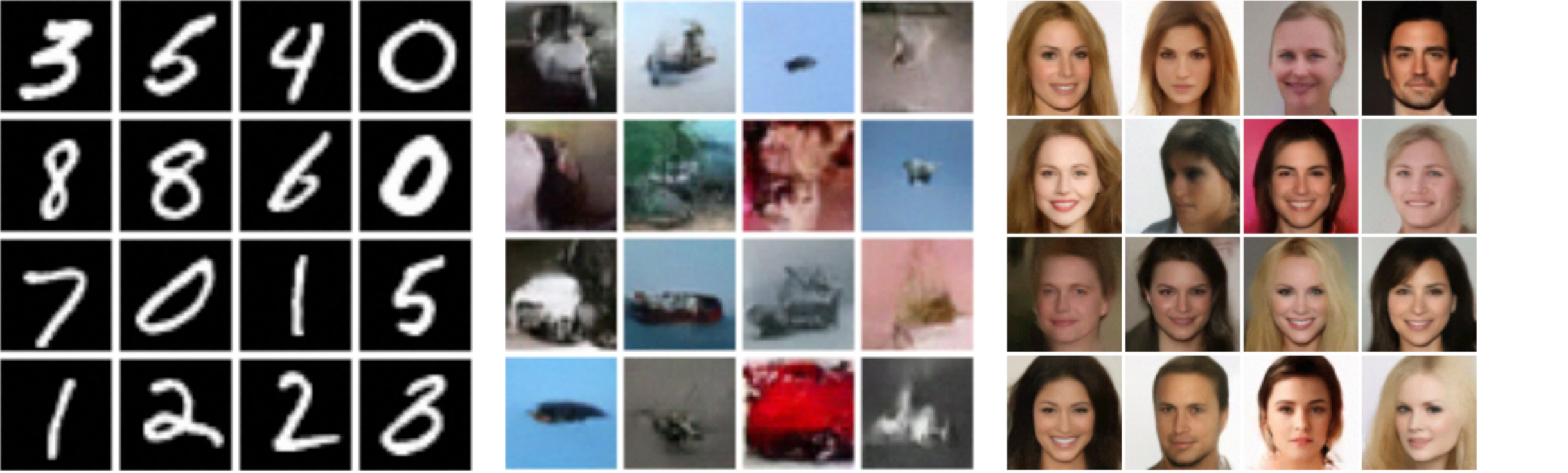}
\caption{Unconditional samples from the base models used for our experiments. From left: MNIST, 5-bit CIFAR-10, and 5-bit CelebA-HQ models.}
\label{fig:unconditional_samples}
\end{figure*}

See \Cref{table:base_parameters} and \Cref{table:pcond_parameters} for the hyperparameters used to define the network architectures train them.
For the color datasets CIFAR-10 and CelebA-HQ, we used 5-bit pixel quantization following \citet{kingma2018glow}. Additionally for CelebA-HQ, we used the same train-test split (27,000/3,000) of \citet{kingma2018glow} and resized the images to $64\times64$ resolution.
Uncurated samples from the base models are included for reference in \Cref{fig:unconditional_samples}.

\begin{table}[!h]
    \centering
    \caption{Hyperparameters used to train the base models used in our experiments.}
    \begin{tabular}{@{}lccc@{}}
        \toprule
        Base Models &  MNIST  & CIFAR-10 & CelebA-HQ \\ \midrule
        Image resolution & $28 \times 28$ & $32 \times 32$ & $64 \times 64$ \\
        Num. scales & 3 & 6 & 6\\
        Res. blocks per scale & 8 & 12 & 10 \\
        Res. block channels & 32 & 64 & 80 \\ \hline
        % Num. parameters & 2607246 & 45055602 & 34582194 \\
        Bits per pixel & 8 & 5 & 5 \\
        Batch size &  128 & 64 & 32 \\
        Learning rate &  $0.001$ & $0.001$ & $0.001$ \\
        % Num. epochs & 200 \\ \hline
        Test set bits-per-dim & 1.053 & 1.725 & 1.268 \\
        \bottomrule
         
    \end{tabular}
    \label{table:base_parameters}
\end{table}
\begin{table}[!h]
\centering
\caption{Hyperparameters used to define and train the pre-generator for each of our experiments.}
\begin{tabular}{@{}lccc@{}}
\toprule
Base Models             &  MNIST         & CIFAR-10         & CelebA-HQ \\ \midrule
Image resolution        & $28 \times 28$ & $32 \times 32$   & $64 \times 64$ \\
Num. scales             & 3              & 4                & 3 \\
Res. blocks per scale   & 3              & 4                & 3 \\
Res. block channels     & 32             & 48               & 48 \\ \hline
Batch size              & 64             & 32               & 8 \\
\end{tabular}
\label{table:pcond_parameters}
\end{table}

\subsection{Hyperparameters: Image Inpainting}
We randomly chose 900/500/300 images from MNIST/CIFAR-10/CelebA-HQ test sets, applied masks defined in \Cref{sec:exp_inpainting}, and generated samples conditioned on the remaining parts.
FID and other sample quality metrics were computed using 6 conditional samples per test image for all MNIST experiments, and 8 conditional samples for all CIFAR-10 and CelebA-HQ experiments. 

\textbf{For VI Methods (Ours \& Ambient VI)}
\begin{itemize}[topsep=1pt,itemsep=0pt,partopsep=0pt,parsep=0pt]
\item  Learning rate: $1\mathrm{e}{-3}$ for MNIST; $5\mathrm{e}{-4}$ for the others
\item  Number of training steps: 4000 for CelebA-HQ; 1000 for the others
\end{itemize}

\textbf{For Langevin Dynamics}
\begin{itemize}[topsep=1pt,itemsep=0pt,partopsep=0pt,parsep=0pt]
\item  Learning rate: $5\mathrm{e}{-4}$ for all datasets
\item  Length of chain: 1000 for CIFAR-10; 4000 for the others
\end{itemize}

\textbf{For PL-MCMC}
\begin{itemize}[topsep=1pt,itemsep=0pt,partopsep=0pt,parsep=0pt]
\item  Learning rate: $5\mathrm{e}{-4}$
\item  Length of chain: 2000 for MNIST
\item $\sigma_a = 1\mathrm{e}{-3}$, $\sigma_p=0.05$
\end{itemize}

\subsection{Hyperparameters: Compressed Sensing}
\textbf{For Ours and \cite{asim2019invertible}}
\begin{itemize}[topsep=1pt,itemsep=0pt,partopsep=0pt,parsep=0pt]
\item Learning rate: $5\mathrm{e}{-4}$
\item Number of training steps: 4000
\item For \cite{asim2019invertible}, we used the same training objective used in their Compressed Sensing experiments: $\argmin_{\bz} \norm{AG(\bz) - \ystar}_2^2$
\end{itemize}

\textbf{For \cite{bora2017compressed}}
\begin{itemize}[topsep=1pt,itemsep=0pt,partopsep=0pt,parsep=0pt]
\item Learning rate: $0.02$
\item Regularization coefficient: $\lambda = 0.1$
\item Following \cite{bora2017compressed}, we repeated each run three times and initialized $\bz_0$ using samples from $\gN(\bzero, \sigma^2\rmI)$ where $\sigma=0.1$. Then we used the best result out of the three runs for evaluation.
\end{itemize}

\subsection{Hyperparameters: Inverse Problems}
Please see \Cref{table:hparams_inverse_problems}.

\begin{table}[!h]
\caption{Hyperparameters for the extra inverse problem experiments.}
\begin{tabular}{@{}ccccc@{}}
\toprule
                & Colorize & CS & CS & SR ($2\times)$ \\ \midrule
Dataset         & \multicolumn{2}{c}{CelebA-HQ}         & \multicolumn{2}{c}{CIFAR-10} \\
Learning rate   & $5\mathrm{e}{-4}$ & $5\mathrm{e}{-4}$ & $5\mathrm{e}{-4}$ & $5\mathrm{e}{-4}$  \\
$\sigma$        & 0.05 & 0.05 & 0.05 & 0.05 \\
Batch size      & 8            & 8                  & 32                 & 32               \\
Number of steps & 1000         & 2000               & 1000               & 1000             \\
\bottomrule
\end{tabular}
\label{table:hparams_inverse_problems}
\end{table}

\end{document}